\documentclass[11pt]{article}
\usepackage{float}
\usepackage{float}
\usepackage[final]{acl}

\usepackage{times}
\usepackage{latexsym}

\usepackage[T1]{fontenc}

\usepackage[utf8]{inputenc}

\usepackage{microtype}

\usepackage{inconsolata}

\usepackage{graphicx}
\usepackage{subcaption}

\usepackage{booktabs}
\usepackage{multirow}

\usepackage[most]{tcolorbox}
\tcbuselibrary{listings,skins,breakable}
\usepackage{xcolor}
\usepackage{listings}
\usepackage{textcomp}

\lstdefinestyle{prompttext}{
    basicstyle=\ttfamily\scriptsize,
    breaklines=true,
    breakatwhitespace=false,
    columns=fullflexible,
    keepspaces=true,
    showstringspaces=false,
    upquote=true,
    escapeinside={(*@}{@*)},
    literate=
      {→}{{$\rightarrow$}}1
      {—}{{---}}1
      {–}{{--}}1
      {“}{{``}}1
      {”}{{''}}1
      {’}{{'}}1
}

\newtcblisting{widepromptbox}[2][]{
    enhanced jigsaw,
    breakable,
    listing only,
    listing engine=listings,
    listing options={style=prompttext},
    colback=gray!3,
    colframe=black!70,
    coltitle=white,
    colbacktitle=black!75,
    title=\textbf{#2},
    fonttitle=\small,
    arc=1.5mm,
    boxrule=0.6pt,
    left=2mm,
    right=2mm,
    top=1mm,
    bottom=1mm,
    width=0.96\textwidth,
    #1
}

\usepackage{graphicx}
\usepackage{booktabs}
\usepackage{multirow}
\usepackage{xcolor}
\usepackage{array}
\newcolumntype{C}[1]{>{\centering\arraybackslash}p{#1}}
\usepackage{tabularx}
\usepackage{xcolor}
\usepackage{amssymb}
\usepackage{colortbl}
\usepackage{hyphenat}
\usepackage{mathtools}
\usepackage{changepage}
\usepackage{enumitem}

\usepackage[most]{tcolorbox}
\definecolor{positive}{HTML}{2E7D32}
\definecolor{negative}{HTML}{B23A48}
\newcommand{\positive}[1]{\textcolor{positive}{\small$\uparrow$\,#1}}
\newcommand{\negative}[1]{\textcolor{negative}{\small$\downarrow$\,#1}}
\newcommand{\posbg}{\cellcolor{positive!12}}
\newcommand{\negbg}{\cellcolor{negative!12}}
\newcolumntype{L}[1]{>{\raggedright\arraybackslash}m{#1}}

\definecolor{must}{RGB}{204,0,0}   
\definecolor{nice}{RGB}{0,102,204} 

\newtcolorbox[list inside=prompt,auto counter]{prompt}[1][]{
    colbacktitle=black!60,
    coltitle=white,
    fontupper=\footnotesize,
    boxsep=5pt,
    left=0pt,
    right=0pt,
    top=0pt,
    bottom=0pt,
    boxrule=1pt,
    #1,
}
\title{\raisebox{-0.6em}{\includegraphics[width=1cm]{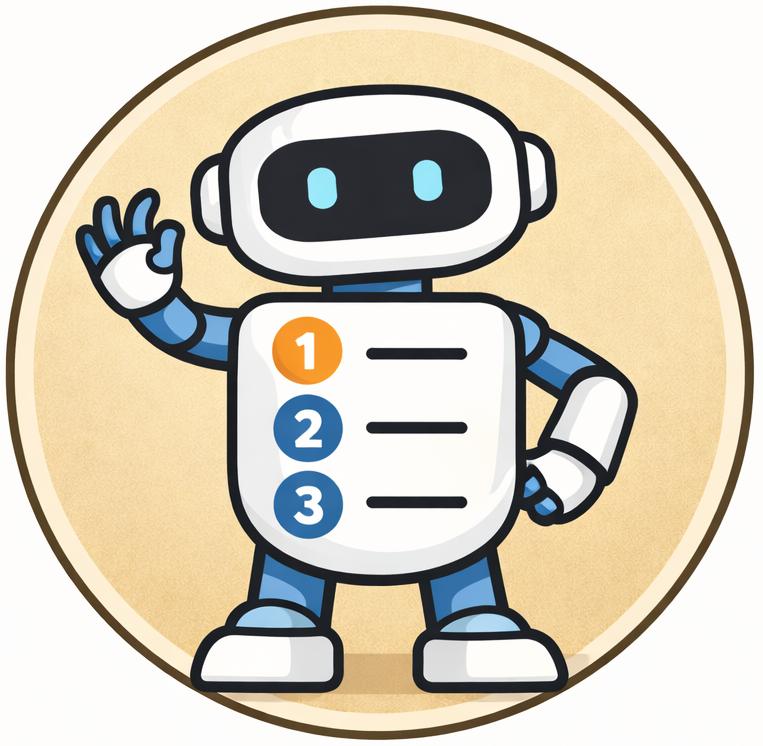}}  First Things First: Teaching MLLM Agents to Prioritize\\ Must-Haves before Nice-to-Haves}

\author{
Tianjie Ju\textsuperscript{1}\thanks{Equal contribution. $^\dag$ Corresponding authors. This work was supported by the National Natural Science Foundation of China (62406188).}\quad
Xinyue Xu\textsuperscript{1,2 *}\quad
Wanxuan Sun\textsuperscript{2}\quad
Lingxiao Diao\textsuperscript{1}\\
\textbf{
Gongshen Liu\textsuperscript{1}\quad
Zhuosheng Zhang\textsuperscript{1 $^\dag$}\quad
Cheng Yang\textsuperscript{2 $^\dag$}
}\\
\textsuperscript{1}School of Computer Science, Shanghai Jiao Tong University\\
\textsuperscript{2}ByteDance\\
\texttt{\{jometeorie, zhangzs\}@sjtu.edu.cn,
yangcheng.iron@bytedance.com}
}

\begin{document}
\maketitle
\begin{abstract}
Recent progress in multimodal large language models (MLLMs) has fueled significant enthusiasm in their potential to act as autonomous agents for real-world tasks. 
However, scenarios requiring agents to fulfill users’ complex, structured requirements remain largely underexplored. 
In this work, we examine reasoning tasks under three distinct requirement scenarios:
(i) Must-have requirements uniquely determine a unique feasible solution; (ii) Multiple answers satisfy the must-have requirements and are prioritized via the nice-to-have requirements; and (iii) No candidate solution satisfies the must-have requirements, in which case the agent should abstain from generating a response. 
We evaluate state-of-the-art MLLMs on carefully constructed problems that reflect realistic service scenarios, including e-commerce, booking, and map-based or ride-hailing. 
Our evaluation reveals that existing MLLMs exhibit catastrophic failures in all scenarios. They frequently misinterpret task requirements, violate must-have requirements, and produce invalid solutions. 
To address this critical gap, we propose First Things First Reinforcement Learning (\textsc{FTF-rl}) that explicitly optimizes reasoning over multi-priority user requirements. 
Experimental results show that our method substantially improves the task success rate compared to strong baselines. 
Moreover, \textsc{FTF-rl} yields general effectiveness on popular logical and mathematical reasoning tasks, including LogicVista, MathVision, and InfoQA.
Our findings suggest that enhancing requirement-aware reasoning capability provides a simple yet effective pathway to improve generalization of MLLM agents. 
Code is available at 
\href{https://github.com/claire62/FTF-RL}{https://github.com/claire62/FTF-RL}.

\end{abstract}

\section{Introduction}

Recent advances in multimodal agents have shown promising results in automating complex tasks such as booking flights, reserving hotels, or navigating graphical user interfaces (GUIs) through natural language instructions~\citep{TravelPlanner, OSWorld, OS-Kairos, OS_GuoYuan}. 
These systems are typically evaluated in settings where users provide unambiguous instructions, 
the agent simply needs to execute the command accurately~\citep{FollowBench, AndroidWorld, CFBench}. 
While impressive, such scenarios represent an idealized setting: the intent is clear, the requirements are minimal, and there is little ambiguity about what constitutes a correct solution. 

In real-world service scenarios, however, user requirements are rarely so simple~\citep{ComplexBench, MPCC, LearnAct}. 
Requests are often multifaceted, containing both \textbf{must-have} requirements 
and \textbf{nice-to-have} requirements.
Current agents often struggle in these settings because they lack a mechanism to prioritize hard requirements over soft requirements, and thus frequently overfit to all expressed conditions, returning either infeasible solutions or incorrect results. 
For example, when asked to ``\textit{book a non-smoking hotel for two people, preferably with breakfast included}'', a typical agent might return a hotel that includes breakfast but fails to ensure the non-smoking requirement, violating the user's core intent. 
Conversely, when given conflicting requirements, such as ``\textit{book the cheapest hotel with two bedrooms, but also with a sea view if possible}'', agents often fail to resolve the trade-off and either refuse to answer or return an irrelevant option.

\begin{figure*}[t!]
  \centering
  \includegraphics[width=0.96\textwidth]{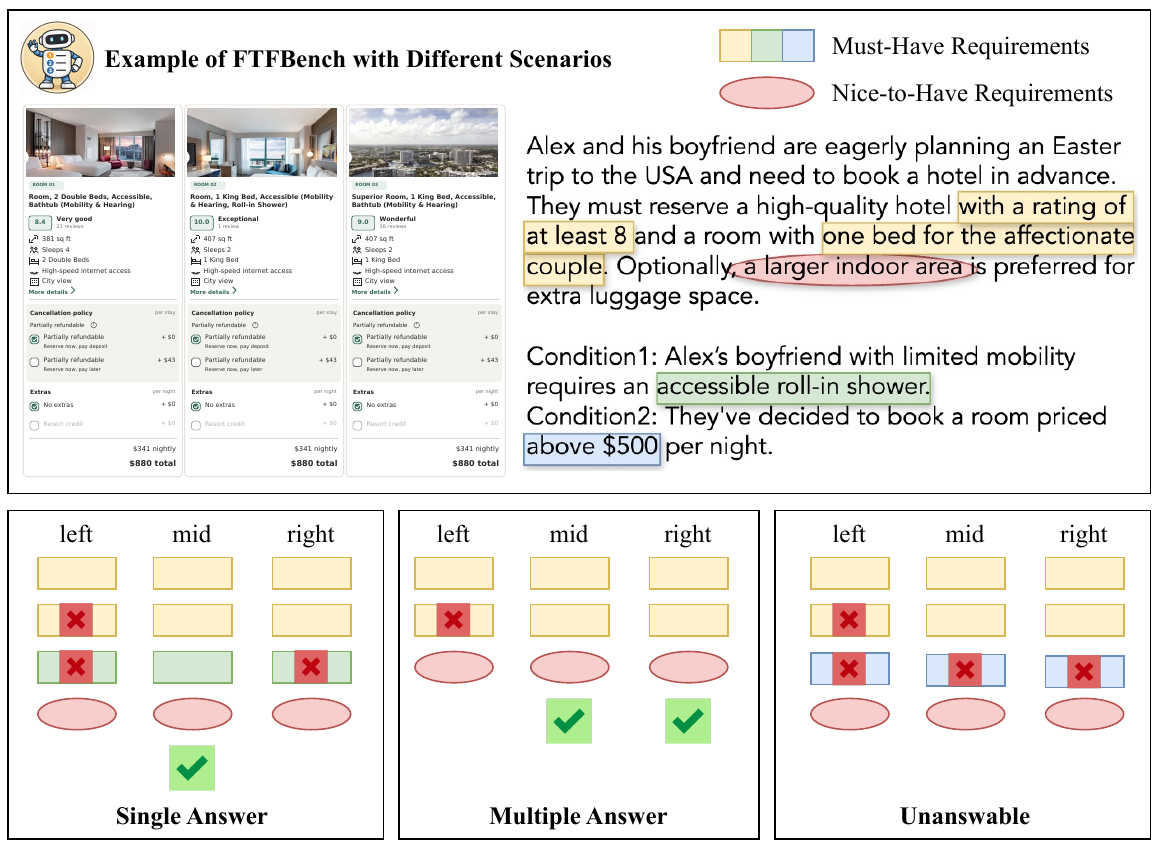}
  \caption{Overview of our proposed \textsc{FTF-bench}, which designed to evaluate requirement-aware reasoning in MLLMs. \textsc{FTF-bench} includes user queries expressed through both must-have and nice-to-have requirements. Tasks are categorized into three settings: must-have requirements with Single Answer (a unique valid candidate satisfies all must-haves), Multiple Answers (several candidates meet the must-haves and must be ranked via nice-to-haves), and Unanswerable (no candidate satisfies all must-haves, requiring abstention).}
  \label{fig: scenarios}
\end{figure*}

To systematically study this problem, we construct the First Things First Benchmark (\textsc{FTF-bench}) to evaluate the capabilities of MLLMs in parsing and reasoning under a clear necessity hierarchy. 
\textsc{FTF-bench} contains more than 3,500 image–requirement pairs from realistic e-commerce, booking, and maps or ride-hailing interfaces. 
For each instance, we synthesize a colloquial user request with must-have and nice-to-have components, verified by human annotators for correctness and clarity. 
All tasks are grouped into three settings that mirror real-world scenarios: 
(i) \textit{must-have requirements} uniquely determine a single valid match; 
(ii) \textit{multiple answers} satisfy all must-have requirements and the system must rank them using soft requirements; 
and (iii) \textit{unanswerable} satisfies the hard requirements (Section~\ref{sec: benchmark}). 

We further introduce a reinforcement learning approach designed to enhance requirement-aware reasoning (\textsc{FTF-rl}). 
Standard SFT~\citep{Self-Instruct, SFT_GPT} and RLHF~\citep{RLHF, RRHF} improve general compliance yet offer little supervision for identifying requirement priority. 
We train MLLMs with a multi-objective reward that checks well-formed outputs, answer correctness, and requirement classification into must-have and nice-to-have, while encouraging intermediate reasoning (Section~\ref{sec: rl}).

We systematically evaluate current MLLMs on \textsc{FTF-bench} and find that catastrophic failures persist widely across domains. 
These MLLMs often over-constrain by elevating requirements to complex rules or under-constrain by ignoring mandatory conditions. 
Errors are most severe in the multiple-answer and unanswerable settings where MLLMs must both respect necessity and resolve trade-offs or abstain. 
When we supply gold requirement labels as an upper-bound setting, accuracy improves substantially, 
which shows that better requirement comprehension alone can substantially unlock the potential of MLLMs (Section~\ref{sec: benchmark results}).

Building on the benchmark diagnosis, we adopt the proposed \textsc{FTF-rl} to train MLLMs for stronger requirement-aware reasoning. 
We apply \textsc{FTF-rl} to Qwen 2.5 VL~\citep{Qwen2.5-VL} of different sizes and observe consistent improvements on \textsc{FTF-bench} across all scenarios. 
Models trained on \textsc{FTF-bench} even show gains on other reasoning benchmarks, indicating that learning to parse and prioritize requirements transfers to broader reasoning skills. 
We call for greater attention to the role of requirement-aware reasoning (Section~\ref{sec: Improve Requirement-Aware Reasoning}).

\section{First Things First Benchmark}
\label{sec: benchmark}
\subsection{Overview}
Real-world intelligent agents, such as customer-service chatbots and virtual assistants, must process user requests involving multiple requirements, where satisfying must-to-haves before nice-to-haves is critical for success. Existing multimodal benchmarks mainly assess generic instruction following and rarely measure whether models can identify and prioritize must-have requirements before optimizing for requirements. To address this gap, we introduce \textsc{FTF-bench}, a comprehensive evaluation suite for assessing MLLMs' ability to understand and reason over complex, multi-priority user requirements. 

\textsc{FTF-bench} covers three representative real-world domains: \textit{e-commerce platforms}, \textit{booking services}, and \textit{maps \& ride-hailing applications} (Figure~\ref{fig: scenarios}). These images span diverse interface layouts and interaction contexts. 
Tasks in \textsc{FTF-bench} are categorized based on the relationship between candidates and mandatory requirements. 

\begin{itemize}[noitemsep, topsep=0pt, leftmargin=1.5em]
    \item Must-have tasks have a unique option that satisfies all mandatory conditions. 
    \item Multi-answer tasks have multiple valid options, requiring models to consider optional requirements and trade-offs to select the best choice. 
    \item Unanswerable tasks contain no valid options, testing whether models can detect infeasible requests and respond appropriately, rather than hallucinating a plausible-looking candidate.
\end{itemize}





\subsection{Task Formulation}
In this section, we formalize the definition of our benchmark tasks, which are designed to evaluate three representative scenarios \emph{unique-answer}, \emph{multiple-answer}, and \emph{unanswerable}. Each sample is represented as:
\begin{equation}
\mathcal{D} = \{x_i\}_{i=1}^{N}, \qquad
x = (I,\mathcal{O},\mathcal{R}^{+},\mathcal{R}^{-}),
\end{equation}
where $I$ is the input image, $\mathcal{O}=\{o_1,\dots,o_M\}$ is the set of $M$ candidate objects appearing in $I$, $\mathcal{R}^{+}$ is the set of must-have requirements, and $\mathcal{R}^{-}$ is the set of nice-to-have requirements. Each requirement $r \in \mathcal{R}^{+}\cup \mathcal{R}^{-}$ is a logical predicate evaluated on a candidate $o \in \mathcal{O}$. The satisfaction of a candidate $o$ with respect to a requirement $r$ is given by:
\begin{equation}
\mathrm{sat}(o,r) \in \{0,1\}, \qquad
\mathrm{sat}(o,r)=1 \Leftrightarrow o \text{ satisfies } r.
\end{equation}


A candidate satisfies all must-haves if
\begin{equation}
\mathrm{sat}^{+}(o) \coloneqq \prod_{r \in \mathcal{R}^{+}} \mathrm{sat}(o,r)=1,
\end{equation}
and the subset of candidates that meet all must-have requirements can be formulated as:
\begin{equation}
\mathcal{S}^{+} = \{o \in \mathcal{O} \mid \mathrm{sat}^{+}(o)=1\}.
\end{equation}

The first scenario, single-answer, is defined by $|\mathcal{S}^{+}|=1$, meaning one candidate satisfies must-have requirements. 

The second scenario, multiple-answer, is defined by $|\mathcal{S}^{+}|\ge 2$. In this case, candidates must be further distinguished using the nice-to-have requirements $\mathcal{R}^{-}$. Each optional requirement $r^{-}_j$ has an associated priority $p(r^{-}_j)\in \mathbb{N}^{+}$. Let $\pi(\mathcal{R}^{-}) = (r^{-}_{(1)}, r^{-}_{(2)}, \dots, r^{-}_{(L)})$ denote the priority-ordered sequence. Starting from $\mathcal{S}_{0}=\mathcal{S}^{+}$, candidates are iteratively filtered according to $\mathcal{S}_{k}=\{o \in \mathcal{S}_{k-1} \mid \mathrm{sat}(o,r^{-}_{(k)})=1\}$. After applying all optional requirements, the remaining candidate set $\mathcal{S}_{L}$ is used to determine the final answer. 

The third scenario, unanswerable, corresponds to $\mathcal{S}^{+}=\varnothing$, meaning no candidate satisfies all must-have requirements. Any prediction of a candidate option in this case is counted as an error.

\subsection{Benchmark Construction}
\label{sec: Benchmark Construction}

\begin{table*}[t!]
\centering
\small
\setlength{\extrarowheight}{3pt}
\begin{tabularx}{\textwidth}{l X X}
\toprule
\textbf{Item} & \textbf{Example 1} & \textbf{Example 2} \\
\midrule
\textbf{Image} 
  & \includegraphics[width=\linewidth]{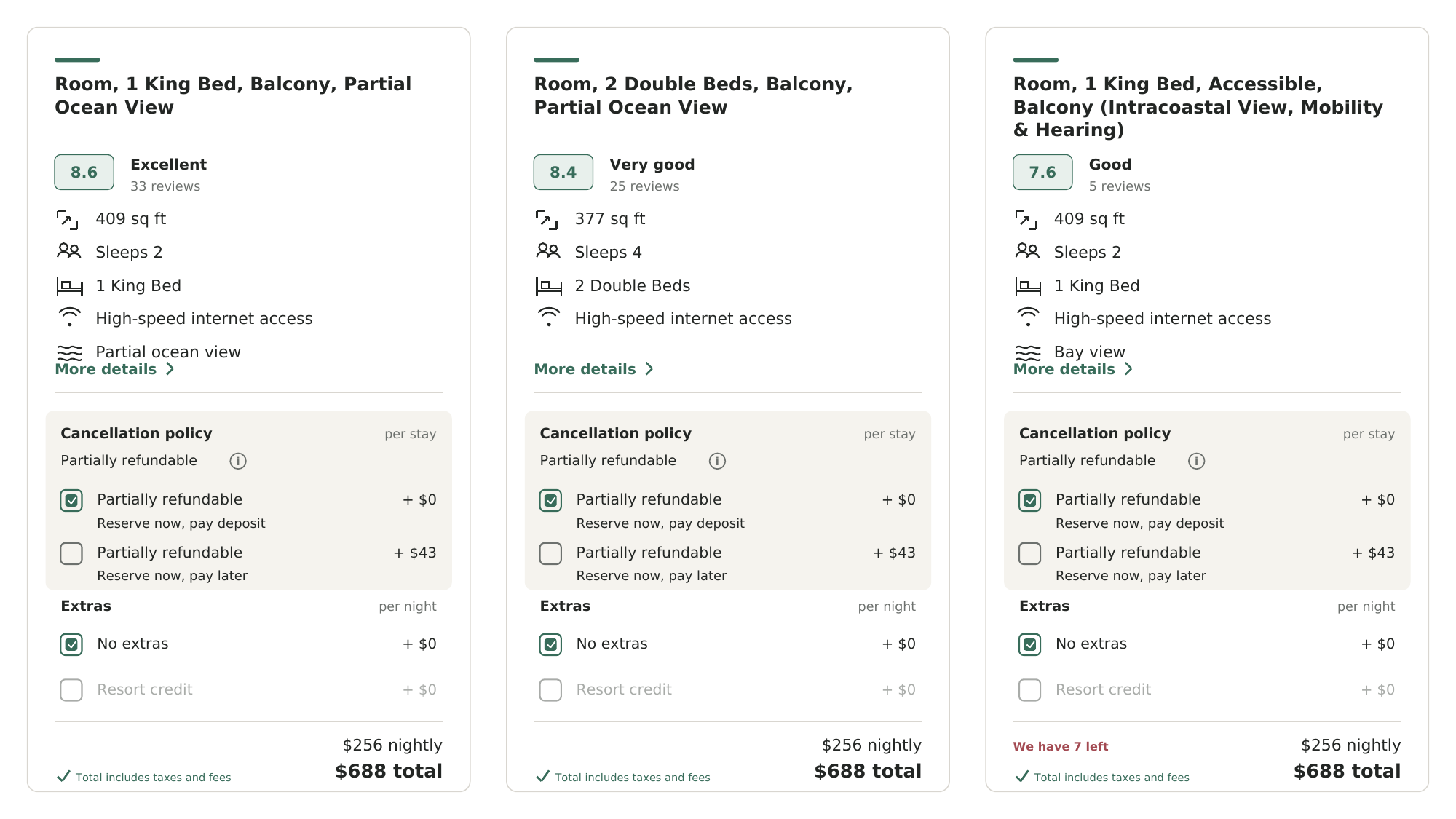} 
  & \includegraphics[width=\linewidth]{figures/hotel.pdf} \\
\midrule
\textbf{Error Type} 
  & Must-have Conflict, No Refusal
  & Misclassification of Nice-to-have as Must-have \\
\midrule
\textbf{User Requirements} 
  & \begin{minipage}[t]{\linewidth}
      \raggedright
      \textcolor{must}{\textbf{Must-have:}} 1 King Bed required.\par
      \textcolor{must}{\textbf{Must-have:}} Must accommodate 4 guests.\par
      \textcolor{nice}{\textbf{Nice-to-have:}} Prefer balcony and partial ocean view.
    \end{minipage}
  & \begin{minipage}[t]{\linewidth}
      \raggedright
      \textcolor{must}{\textbf{Must-have:}} Price per night \(\leq\) \$300.\par
      \textcolor{must}{\textbf{Must-have:}} Must accommodate 2 guests.\par
      \textcolor{nice}{\textbf{Nice-to-have:}} Prefer rating \(\geq\) 8.5.
    \end{minipage} \\
\midrule
\textbf{Model Output (Incorrect)}
  & Recommended “Room, 2 Double Beds, Balcony, Partial Ocean View”, 377 sq ft, accommodates 4 guests, includes balcony and partial ocean view — claimed to fully meet the request.
  & “Sorry, no available rooms meet the requirements.” \\
\midrule
\textbf{Expected Output}
  & The system should reject or indicate no available room meets all must-have requirements.  
    Example: “No room satisfies both the ‘1 King Bed’ and ‘4 guests’ requirements.”
  & The system should treat rating criterion as optional and list all rooms satisfying the price and capacity requirements, possibly ranked by rating.  
    Example: “3 rooms found under \$300 for 2 guests, sorted by rating.” \\
\bottomrule
\end{tabularx}
\caption{Comparison of two error examples, presented in a column-wise layout for side-by-side inspection.}
\label{tab:rotated_error_examples}
\end{table*}

To construct a high-quality dataset for training and evaluation, we develop a large-model-driven pipeline that systematically generates image-requirement pairs. The pipeline comprises the following core stages: Image Collection, Requirement Generation, Colloquial Requirement Expression, and Human Verification. 
Table~\ref{tab:rotated_error_examples} presents representative instances from \textsc{FTF-bench}, illustrating typical error types made by MLLMs. It shows how models often misclassify must-have versus nice-to-have requirements or fail to properly reject infeasible requests.

\paragraph{Image Collection.} We first assemble a diverse image corpus to capture realistic service-oriented scenarios, covering common user-facing applications in daily life. For each domain, volunteers collect screenshots from multiple applications, sampling different interface states (e.g.\ homepages, product pages, shopping carts). 

\paragraph{Requirement Generation.} For each image, we leverage LLMs to generate both requirements and the corresponding ground-truth answers. We design prompt templates to align precisely with three task types (single-answer, multi-answer, unanswerable), ensuring that each requirement is grounded in the image content. 
We produce a labeled requirement set that separates must-have ($\mathcal R^{+}$) from nice-to-have ($\mathcal R^{-}$) requirements, and for multi-answer cases we assign a priority order over $\mathcal R^{-}$ to support ranking among candidates, thereby explicitly modeling trade-offs and partial satisfaction.

\paragraph{Colloquial Requirement Expression.} We then convert structured requirements into natural, conversational queries, better emulating real user behavior. In practice, users rarely frame needs in formal subclauses. They pose a single coherent question that interleaves must-have and nice-to-have requirements. To simulate this, we prompt the model to rewrite requirements as context-aware utterances while preserving priority hierarchy. 
This step preserves the explicit must-have/nice-to-have distinction in a single utterance, so that downstream evaluation can still reflect the necessity-first objective rather than generic instruction-following.

\paragraph{Human Verification.} Finally, we apply human verification to secure benchmark integrity. Four trained annotators independently inspect each image–query pair, validating requirement correctness, answer consistency, and the separation between mandatory and optional constraints. Any disagreement is resolved via discussion, producing a high-quality, standardized dataset that supports robust evaluation. 
During checking, annotators ensure that each instance conforms to one of the three settings so that the final benchmark faithfully measures requirement-aware reasoning. 
We provide several prominent error types in Appendix~\ref{sec: Details of Human Verification}.

\begin{figure*}[t!]
  \centering
  \includegraphics[width=1\textwidth]{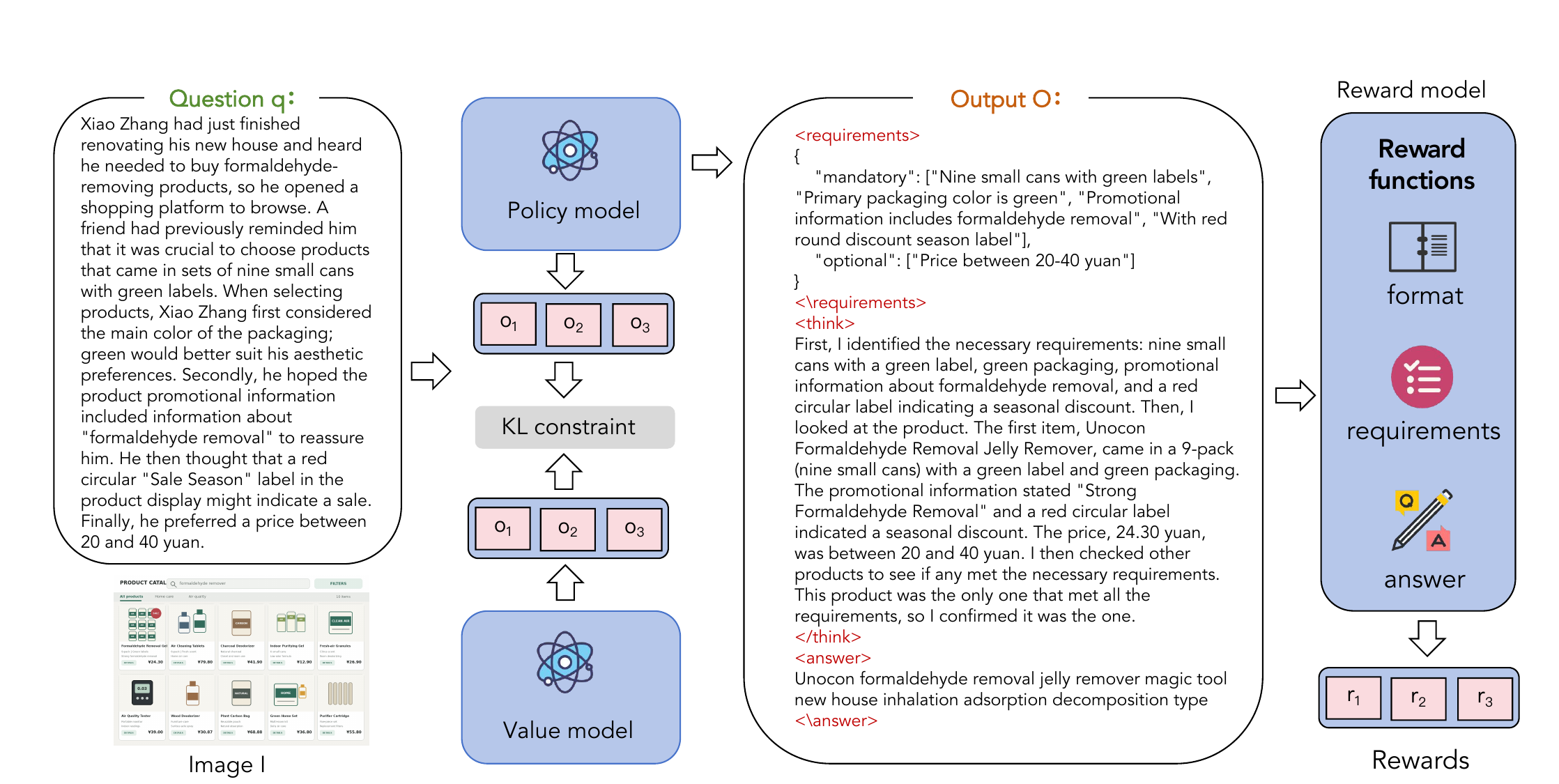}
  \caption{\textbf{\textsc{FTF-rl} framework.} Given an input image $I$ and question $q$, we first sample $G$ candidate outputs $\{o_i\}_{x wi=1}^{G}$ from the old policy model. Then we compute a reward $r_i$ for each $o_i$ using our proposed multi-objective reward function (see Section~\ref{sec: Multi-objective Reward Functions} for details). Finally, we optimize the current policy model by maximizing $A_i$, while regularizing the update using KL divergence between $\pi_{\theta}$ and the reference policy model $\pi_{\text{ref}}$ to keep the updated policy close to the reference policy.}
  \label{fig: ftf-rl-framework}
\end{figure*}

\section{First Things First Reinforcement Learning}
\label{sec: rl}

\subsection{Multi-objective Reward Functions}
\label{sec: Multi-objective Reward Functions}
Reward models play a crucial role in RL, as they directly determine the optimization signal that guides policy improvement. Recent advances, such as DeepSeek-R1~\citep{DeepSeek-R1}, have demonstrated that verifiable reward functions can substantially enhance the reasoning ability of MLLMs. Inspired by this success, we design a rule-based, multi-objective reward function to evaluate both requirement understanding and visual reasoning capabilities. This design ensures that the model not only produces correct final answers but also generates interpretable intermediate steps, thereby improving generalization. 

Specifically, our reward framework assesses model outputs along four key dimensions: format compliance, final-answer correctness and requirement classification accuracy, which distinguishes between must-have and nice-to-have requirements. Figure~\ref{fig: ftf-rl-framework} provides an overview of the framework.

\textbf{Format Reward.} 
The format reward $R_{\text{format}}$ verifies whether the model output strictly follows the predefined XML-style schema. 
Specifically, we check three components: (i) whether the requirement classification is enclosed within \texttt{<requirements>} \ldots \texttt{</requirements>} tags and expressed as a valid JSON object separating \texttt{must\_have} and \texttt{nice\_to\_have} fields; 
(ii) whether the intermediate reasoning process is included in \texttt{<think>} \ldots \texttt{</think>} tags; 
and (iii) whether the final answer is provided in \texttt{<answer>} \ldots \texttt{</answer>} tags. 
The reward is computed as follows: 
\begin{equation}
R_{\text{format}} = 
\begin{cases}
1, & \text{if all components are well-formed}, \\
0, & \text{if no component follows the format}.
\end{cases}
\end{equation}


\textbf{Accuracy Reward.} 
The accuracy reward $R_{\text{answer}}$ measures whether the model's final prediction matches the ground-truth answer. 
Since our tasks are formulated as fill-in-the-blank questions, we employ a powerful MLLM as a judging model to robustly compare the model-generated answer against the reference solution. 
This approach mitigates surface-form mismatch issues (e.g., synonyms, equivalent expressions). 
Formally, for a given question $q$, model output $y$, and ground-truth answer $a^{*}$, we define:

\begin{equation}
R_{\text{answer}} = 
\begin{cases}
1, & \text{Judger}(q, \text{Ans}(y), a^{*}) = \texttt{True}, \\
0, & \text{otherwise},
\end{cases}
\end{equation}
where $\text{Ans}(y)$ extracts the content between the \texttt{<answer>} \ldots \texttt{</answer>} tags from the model output, 
and $\text{Judger}(\cdot)$ is the MLLM-based evaluation function that returns \texttt{True} if the predicted answer is semantically equivalent to the ground-truth $a^{*}$.

\textbf{Requirement Reward.} 
The requirement reward $R_{\text{requirement}}$ evaluates the model’s ability to correctly identify and classify user requirements into \texttt{must\_have} and \texttt{nice\_to\_have} categories. 
This component is motivated by the observation that many reasoning failures arise not from incorrect computation, but from misinterpretation of user intent: 
models often over-constrain by treating nice-to-have requirements as hard requirements, or under-constrain by ignoring must-have requirements, leading to invalid or suboptimal answers. 
By explicitly rewarding correct requirement classification, we encourage the model to faithfully represent user intent before performing reasoning, thereby reducing downstream reasoning errors and improving decision quality. 

We utilize the Macro-averaged $F_1$ score as the primary metric to ensure balanced optimization across both categories. Formally, let $Y = \{y_1, \dots, y_n\}$ be the ground truth labels for a set of requirements and $\hat{Y} = \{\hat{y}_1, \dots, \hat{y}_n\}$ be the predicted labels. The reward $R_{\text{requirement}}$ assigned to a generated response is defined as:
\begin{equation}
R_{\text{requirement}} = \alpha \cdot \text{Macro-}F_1(Y, \hat{Y}),
\end{equation}
where $\text{Macro-}F_1(Y, \hat{Y})$ represents the arithmetic mean of the class-specific $F_1$ scores:
\begin{equation}
\text{Macro-}F_1 = \frac{1}{|\mathcal{C}|} \sum_{c \in \mathcal{C}} F_1(Y_c, \hat{Y}_c),
\end{equation}
where $\mathcal{C} = \{\text{must\_have, nice\_to\_have}\}$ is the set of requirement classes. 
This design provides a soft, differentiable reward signal that penalizes both false positives (e.g., over-constraining) and false negatives (e.g., ignoring essential requirements), thereby promoting balanced optimization.

\section{Experiments}

\begin{table*}[t!]
\centering
\renewcommand{\arraystretch}{1.2}
\resizebox{0.85\linewidth}{!}
{%
\begin{tabular}{lcc|cc|cc|cc}
\toprule
\textbf{Models} 
& \multicolumn{2}{c|}{\textbf{Single-Answer}}
& \multicolumn{2}{c|}{\textbf{Multiple-Answer}}
& \multicolumn{2}{c|}{\textbf{Unanswerable}}
& \multicolumn{2}{c}{\textbf{Average}} \\
\cmidrule(lr){2-3}\cmidrule(lr){4-5}\cmidrule(lr){6-7}\cmidrule(lr){8-9}
& \textbf{Upper} & \textbf{Direct}
& \textbf{Upper} & \textbf{Direct}
& \textbf{Upper} & \textbf{Direct}
& \textbf{Upper} & \textbf{Direct} \\
\midrule
\multicolumn{9}{c}{\textit{Proprietary MLLMs}} \\
\midrule
Gemini-2.5-pro{\textsuperscript{\dag}} &  \textbf{88.89} & \negbg 86.91 & \textbf{84.20} & \negbg 82.26 & \textbf{81.72} & \negbg 78.55 & \textbf{84.26} & \negbg 81.75 \\
GPT-5{\textsuperscript{\dag}} &  83.57 & \posbg \textbf{85.21} & \textbf{80.73} & \negbg 78.10 & \textbf{82.95} & \negbg 80.24 & \textbf{82.49} & \negbg 80.86 \\
GPT-o3{\textsuperscript{\dag}} & \textbf{82.67} & \negbg 77.78 & \textbf{80.41} & \negbg 80.03 & \textbf{83.31} & \negbg 83.09 & \textbf{82.33} & \negbg 79.68 \\
Doubao-1.6-seed{\textsuperscript{\dag}} & \textbf{80.09} & \negbg 78.66 & \textbf{81.70} & \negbg 76.99 & 83.80 & \posbg \textbf{84.05} & \textbf{82.24} & \negbg 80.59 \\
\midrule
\multicolumn{9}{c}{\textit{Open-Source MLLMs}} \\
\midrule
LLaMA-4 & \textbf{58.36} & \negbg 55.48 & \textbf{55.93} & \negbg 54.45 & \textbf{44.43} & \negbg 36.94 & \textbf{55.03} & \negbg 51.99 \\
Qwen2.5-VL-7B-Instruct & \textbf{61.16} & \negbg 22.99 & \textbf{58.25} & \negbg 20.10 & \textbf{47.59} & \negbg 18.64 & \textbf{57.69} & \negbg 21.05 \\
Qwen2.5-VL-32B-Instruct & \textbf{70.66} & \negbg 69.25 & \textbf{70.55} & \negbg 70.17  & \textbf{53.08} & \negbg 50.58 & \textbf{67.72} & \negbg 66.57 \\
Qwen2.5-VL-72B-Instruct & \textbf{71.26} & \negbg 35.76 & \textbf{74.75} & \negbg 35.12 & \textbf{83.09} & \negbg 29.62 & \textbf{77.60} & \negbg 34.48 \\
{LLaVA-OneVision-7B} & {\textbf{45.28}} & {\negbg 19.15} & {\textbf{43.17}} & {\negbg 18.22} & {\textbf{42.36}} & {\negbg 21.78} & {\textbf{43.38}} & {\negbg 19.73} \\
{LLaVA-NEXT-34B} & {\textbf{50.31}} & {\negbg 27.27} & {\textbf{50.20}} & {\negbg 20.21} & {\textbf{51.40}} & {\negbg 25.83} & {\textbf{50.76}} & {\negbg 24.11} \\
{LLaVA-NeXT-13B} & {\textbf{48.65}} & {\negbg 26.92} & {\textbf{46.89}} & {\negbg 18.34} & {\textbf{46.72}} & {\negbg 33.95} & {\textbf{47.25}} & {\negbg 26.13} \\
{LLaVA-1.5-13B} & {\textbf{48.70}} & {\negbg 26.98} & {\textbf{46.94}} & {\negbg 16.39} & {\textbf{42.77}} & {\negbg 34.00} & {\textbf{45.72}} & {\negbg 25.28} \\
\bottomrule
\end{tabular}%
}
\caption{Evaluation results of current MLLMs on \textsc{FTF-bench} across three task types. Upper denotes the performance when MLLMs are given golden requirements, while Direct reports accuracy when MLLMs directly answer without additional guidance.}
\label{tab:eval}
\end{table*}

\label{sec: rl results}
\begin{table*}[t!]
\centering
\resizebox{0.85\linewidth}{!}
{
\begin{tabular}{l|ccc|c}
\toprule
\textbf{Model} & \textbf{Single-Answer} & \textbf{Multiple-Answer} & \textbf{Unanswerable} & \textbf{Average} \\
\midrule
Qwen2.5-VL-3B-Instruct & 48.55 & 33.13 & 32.79 & 38.95 \\
\quad + \textsc{FTF-rl} & \textbf{56.52}\positive{7.97} & \textbf{41.10}\positive{7.97}  & \textbf{34.42}\positive{1.63} & \textbf{45.85}\positive{6.90} \\
\midrule
Qwen2.5-VL-7B-Instruct & 46.38 & 32.52 & \textbf{44.26}  & 39.78 \\
\quad + \textsc{FTF-rl} & \textbf{57.97}\positive{11.59} & \textbf{58.90}\positive{26.38} & 42.62\negative{1.64} & \textbf{55.80}\positive{16.02} \\
\midrule
{LLaVA-OneVision-7B} & {45.82} & {33.69} & {42.87} & {39.86} \\
{\quad + \textsc{FTF-rl}} & {\textbf{52.51}\positive{6.69}} & {\textbf{40.15}\positive{6.46}} & {\textbf{48.79}\positive{5.92}} & {\textbf{46.32}\positive{6.46}} \\
\midrule
{LLaVA-1.5-13B }& {49.23} & {37.51} & {43.19} & {42.93} \\
{\quad + \textsc{FTF-rl}} & {\textbf{57.81}\positive{8.58}} & {\textbf{43.74}\positive{6.23}} & {\textbf{49.05}\positive{5.86}} & {\textbf{50.00}\positive{7.07}} \\
\bottomrule
\end{tabular}}
\caption{Impact of the proposed reinforcement learning method on requirement-aware reasoning.}
\label{tab: rl}
\end{table*}

\subsection{Setup}
\label{sec: Setup}
We evaluate seven widely used MLLMs on \textsc{FTF-bench}. 
The proprietary group includes Gemini 2.5 Pro~\citep{Gemini-2.5}, GPT-5~\citep{GPT-5}, GPT-o3~\citep{GPT-o3}, and Doubao 1.6 Seed~\citep{Seed1.6}.
The open-source group covers Qwen2.5-VL~\citep{Qwen2.5-VL}, LLaMA-4~\citep{LLaMA-4}, and four variants of LLaVA~\citep{LLaVA}. 
For Qwen2.5-VL, we report results for the 7B, 32B, and 72B checkpoints to examine scaling effects. 

We conduct experiments on all instances of \textsc{FTF-bench}. 
Detailed prompts used in both Direct and Upper settings are provided in Appendix~\ref{sec: Evaluation Prompts}. 
For reinforcement learning, we train MLLMs using 90\% of the benchmark data and reserve the remaining 10\% for evaluation. 
All reinforcement learning experiments are conducted under a unified evaluation to achieve fair comparison. 
We adopt GRPO~\citep{GRPO} with a KL-penalty coefficient of $10^{-2}$ to stabilize policy updates. Training is conducted with a global batch size of $128$ and a learning rate of $10^{-6}$, using AdamW optimization and full-shard FSDP for efficient distributed training. During rollout, we use a temperature of $1.0$ and sample $n=5$ candidate responses.

\subsection{Evaluation results on \textsc{FTF-bench}}
\label{sec: benchmark results}


We conduct experiments of current MLLMs on \textsc{FTF-bench} under two input settings. 
In the Direct setting, MLLMs read the original colloquial user request and must infer both the requirement hierarchy and the final decision. 
In the Upper setting, we paraphrase the same request into gold requirement labels that separate must-haves from nice-to-haves and feed these labels to MLLMs as structured guidance, which serves as an upper bound for performance with perfect requirement understanding. 
The main results are shown in Table~\ref{tab:eval}.

\textbf{Upper exceeds Direct across most scenarios, confirming that the primary source of error is not visual perception alone but the failure to correctly parse and prioritize requirements from natural language}. 
Nearly all MLLMs show a noticeable decline when directly interpreting the original user text. 
The gap is particularly severe in the multiple-answer and unanswerable scenarios. 
For multiple-answer tasks, candidate options often contain several plausible distractors, and only by reasoning with nice-to-have conditions can the optimal choice be determined. 
Without explicit requirement decomposition, MLLMs struggle to resolve these fine-grained trade-offs. 
For unanswerable tasks, where no candidate satisfies the must-have requirements, MLLMs should reject all options. 
However, most MLLMs still attempt to produce an answer, revealing a tendency to over-accommodate user prompts even when abstention is the correct strategy. 
These observations expose a fundamental weakness in requirement-aware reasoning.

\textbf{Another observation is the large gap between model performance and the upper bound, especially for open-source MLLMs.}
Qwen2.5-VL and LLaMA-4 exhibit substantial improvements once gold requirements are provided, suggesting that their reasoning pipeline is hampered less by intrinsic visual limitations than by misinterpretation of complex user intent.

Interestingly, scaling effects are not monotonic. 
Within Qwen2.5-VL, the 7B and 72B models fail catastrophically in direct reasoning, whereas the 32B variant performs better. 
We interpret this as evidence of two distinct failure modes. 
The smaller 7B model appears to lack the basic capacity to reliably distinguish must-haves from nice-to-haves, which leads to frequent violations of core requirements. 
The largest 72B model surprisingly shows an opposite failure pattern. 
Its tendency to overfit to the surface form of user prompts results in excessive alignment with every expressed condition, elevating optional requirements to mandatory status and thereby producing infeasible outputs. 
In contrast, the 32B model achieves a more balanced handling of requirement prioritization, revealing that scale alone does not guarantee progress.


\subsection{Improve Requirement-Aware Reasoning}
\label{sec: Improve Requirement-Aware Reasoning}

\subsubsection{Main results}
We further improve requirement-aware reasoning by applying \textsc{FTF-rl} to the open-source Qwen2.5-VL. 
We randomly sample 90\% of \textsc{FTF-bench} for training and hold out the remaining 10\% for evaluation. 
Comparison results before and after applying \textsc{FTF-rl} are presented in Table~\ref{tab: rl}, which evaluates models on the held-out 10\% subset, with Average computed over all instances in this subset rather than as the mean of the three scenario-level scores. 
We have further verified that this 10\% evaluation subset is representative of the full benchmark in Appendix~\ref{app:subset-analysis}.

\textbf{Both the 7B and 3B variants deliver strong gains, demonstrating that \textsc{FTF-RL} is highly effective at strengthening the requirement-aware reasoning capabilities of MLLMs.}
The most remarkable results appear in the multiple-answer scenario. 
Qwen2.5-VL-7B-Instruct improves by more than 26\% after reinforcement learning, showing that the proposed \textsc{FTF-RL} enables MLLMs to reliably separate must-haves from nice-to-haves and make the correct choice even when several confusing candidates are present. 
The single-answer and unanswerable settings also benefit, with consistent improvements observed across scales.

\subsubsection{Generalization across Reasoning Benchmarks}

To further investigate how requirement-aware reasoning contributes to the general capability of MLLMs, we compare MLLMs trained only on \textsc{FTF-bench} with their baselines on other logic and math reasoning benchmarks, including LogicVista~\citep{LogicVista}, MathVision~\citep{MathVision}, and InfoQA~\citep{InfoQA}. 
Table~\ref{tab:Generality} presents the results of Qwen2.5-VL models before and after \textsc{FTF-rl} training. 
We are surprised to find that even without any explicit training on these reasoning benchmarks, the MLLMs exhibit consistent improvements after reinforcement learning on most reasoning benchmarks. 

This suggests that \textbf{requirement-aware reasoning not only strengthens the understanding of complex user intent but also stimulates the general reasoning ability of MLLMs, leading to clear gains across diverse and challenging multimodal reasoning tasks}. 
Two case studies on LogicVista and MathVision are provided in Appendix~\ref{app:qualitative_reasoning}. 
Ablation studies on reward functions are provided in Appendix~\ref{appendix: Ablation study of reward functions}.


\begin{table}[t!]
\centering
\renewcommand{\arraystretch}{1.2}
\resizebox{0.98\linewidth}{!}{
\begin{tabular}{l|cccc}
\toprule
\textbf{Model} &  \textbf{LogicVista} & \textbf{MathVision} & \textbf{{InfoQA}} \\
\midrule
Qwen2.5-VL-7B-Instruct &  43.40 & 24.67 & {{65.35}}\\
\quad + \textsc{FTF-rl} & \textbf{47.43}\positive{4.03} & \textbf{25.65}\positive{0.98} & {\textbf{67.81}\positive{2.46}} \\
\midrule
Qwen2.5-VL-3B-Instruct & 36.91 & 23.03 & {{37.62}} \\
\quad + \textsc{FTF-rl} & \textbf{40.49}\positive{3.58} & \textbf{24.34}\positive{1.31} & {\textbf{39.93}\positive{2.31}} \\
\midrule
{LLaVA-1.5-13B} & {29.23} & {11.12} & {{41.57}} \\
{\quad + \textsc{FTF-rl}} & {\textbf{35.61}}\positive{6.38} & {\textbf{13.74}}\positive{2.62} & {\textbf{41.92}\positive{0.35}}  \\
\bottomrule
\end{tabular}}
\caption{Effect of \textsc{FTF-rl} on model generalization across various reasoning benchmarks.}
\label{tab:Generality}
\end{table}

\subsubsection{Ablation study of reward functions}
\label{appendix: Ablation study of reward functions}
To evaluate the contribution of requirement classification supervision, we further conduct an ablation study by removing $R_{\text{requirement}}, R_{\text{format}}, R_{\text{answer}}$ from the overall objective and retraining the model under identical settings, respectively. 
The ablation results across \textsc{FTF-rl} and three other reasoning benchmarks are summarized in Table~\ref{tab:ablation}. 
We observe a consistent performance drop across all datasets. 
The performance drop indicates that all reward components contribute to the final performance. 
This finding confirms that multi-objective reinforcement learning, which jointly optimizes for format compliance, answer accuracy, requirement understanding, and reasoning quality, is crucial for improving accuracy in multimodal document understanding tasks.

\begin{table*}[t!]
\centering
\resizebox{0.8\linewidth}{!}{
\begin{tabular}{l|c|ccc}
\toprule
\textbf{Setting} & \textbf{\textsc{FTF-bench}\textsubscript{val}} & \textbf{LogicVista} & \textbf{MathVision} & \textbf{InfoQA} \\
\midrule
\textsc{FTF-rl} & 55.8 & 47.4 & 25.7 & 67.8 \\
w/o $R_{\text{requirement}}$ & 52.4\negative{3.4} & 44.7\negative{2.7} & 24.7\negative{1.0} & 60.3\negative{7.5} \\
w/o $R_{\text{format}}$ & {54.5\negative{1.3}} & {45.8\negative{1.6}} & {25.1\negative{0.6}} & {64.5\negative{3.3}} \\
w/o $R_{\text{answer}}$ & {48.3\negative{7.5}} & {43.7\negative{3.7}} & {24.5\negative{1.2}} & {61.2\negative{6.6}} \\
\bottomrule
\end{tabular}}
\caption{{Ablation study of the different rewards on Qwen2.5-VL-7B-Instruct.}}
\label{tab:ablation}
\end{table*}

\section{Related Work}
\label{gen_inst}

\subsection{Multimodal Reasoning}
Research on reasoning in MLLMs has advanced through Multimodal Chain-of-Thought (MCoT) and RL methods~\citep{junlin2025large}. 
To inject more procedural structure, early CoT-based approaches~\citep{DBLP:conf/naacl/LiLZQHW25} decomposed reasoning into stages such as perceptual summarization, localized grounding, and fine-grained verification. However, their rigid pipeline structure limited adaptability to diverse tasks.  
Subsequent variants adopted more flexible decomposition of reasoning: for example, Cantor~\citep{DBLP:conf/mm/GaoCZFSZZZSCJ24} explicitly partitions the model’s processing into perception and decision steps, while TextCoT~\citep{DBLP:journals/corr/abs-2404-09797} refines reasoning by zooming from global descriptions to local crop analyses.  
For specialized reasoning, \citet{DBLP:journals/corr/abs-2501-03230} proposed Video-of-Thought (VoT), which breaks down video clips into temporally ordered segments to improve action prediction within video-chat benchmarks.

Reinforcement learning has emerged to elicit deeper reflection and optimize reasoning quality. RL frames reasoning as a Markov Decision Process (MDP), optimizing trajectories via rewards. \citet{DBLP:journals/corr/abs-2501-12948} used verifiable rewards (e.g., math correctness), which \citet{DBLP:journals/corr/abs-2502-19634} extended to MLLMs with MedVLM-R1. For spatial reasoning, \citet{DBLP:journals/corr/abs-2504-01805} used spatial consistency rewards to boost 6D reasoning accuracy. Recent RL advances like StepGRPO added intermediate rewards, enhancing logical consistency on R1-VL~\citep{DBLP:journals/corr/abs-2503-12937}. However, most prior RL approaches focus on pushing the upper bound of VLM reasoning performance, without explicitly ensuring that the model fully understands and decomposes the requirements of the question. 

\subsection{Instruction Following Benchmarks in MLLMs}
A variety of benchmarks have been introduced to evaluate multimodal large language models (MLLMs) across multiple facets~\citep{zhang2025large}, including general knowledge~\citep{DBLP:conf/eccv/LiuDZLZZYWHLCL24, DBLP:conf/cvpr/YueNZ0LZSJRSWYY24}, document understanding~\citep{DBLP:conf/cvpr/YueNZ0LZSJRSWYY24, DBLP:journals/corr/abs-2306-13394}, perceptual reasoning~\citep{DBLP:conf/cvpr/HuSDR20}, multi-image comprehension~\citep{DBLP:conf/icml/YuYLWL0WW24}, and instruction following~\citep{DBLP:conf/iclr/QianYFGYG25, DBLP:conf/nips/BittonBHSZAGTS23}.  

Several benchmarks have been proposed to quantify the instruction-following capability of large language models~\citep{DBLP:conf/acl/Jiang0ZZLMS00W24, DBLP:conf/acl/ZhangZSLZLYLQC025, DBLP:journals/corr/abs-2406-16356}. For instance, LIFBench~\citep{DBLP:conf/acl/WuWLSYLZ025} limits its scope to instructions whose fulfillment can be verified automatically, which enhances model differentiation.  
InfoBench ~\citep{DBLP:conf/acl/QinSHYCWW00Y24} decomposes prompts into sub-instructions and computes the Decomposed Requirements Following Ratio (DRFR) to grant partial credit for each satisfied micro-requirement.  
Benchmarks like MIA-Bench~\citep{DBLP:conf/iclr/QianYFGYG25} and VisIT-Bench~\citep{DBLP:conf/nips/BittonBHSZAGTS23} adopt GPT-4 for question generation and evaluation in instruction-following settings.  
However, existing instruction-following benchmarks implicitly treat all directives as equally important.  

\section{Conclusion}

In this paper, we investigate requirement-aware reasoning in realistic service settings. 
We first introduce \textsc{FTF-bench} to evaluate the requirement-aware reasoning capability across various scenarios. 
Experiments show that current MLLMs frequently misread requirements, violate hard requirements, and output invalid solutions, leading to catastrophic failures in prioritizing requirements. 
To improve the requirement-aware reasoning capability, we further present a multiobjective RL framework \textsc{FTF-rl} that rewards correct requirement identification, proper ordering of requirement satisfaction, and answer validity. 
After training, MLLMs perform substantially better and narrow the gap toward the upper bound obtained with requirement labels. 
We also observe consistent gains on other tasks that require complicated reasoning, which suggests that strengthening requirement comprehension yields broader generalization. 
We call for more attention to the pivotal role of requirement-aware reasoning in advancing the reasoning reliability of MLLMs.

\section*{Limitations}

Although \textsc{FTF-bench} covers representative service-oriented scenarios, including e-commerce, booking, and map-based applications, it does not yet exhaust the broader space of real-world requirement-aware decision making. 
Future extensions could incorporate more diverse domains where user requirements may evolve over time. 
In addition, our reinforcement learning experiments are conducted on the current scale of \textsc{FTF-bench}. 
While the observed gains on both in-domain and out-of-domain reasoning benchmarks suggest that requirement-aware reasoning can improve the generalization of MLLMs, scaling the training data further would help better characterize the upper limit of this effect and test how far requirement-aware supervision can transfer across broader multimodal reasoning tasks.

\section*{Ethical Considerations}

Our study focuses on evaluating and enhancing requirement-aware reasoning in MLLMs. 
We call for the widespread deficiencies of current models in prioritizing must-have over nice-to-have requirements and introduce reinforcement learning methods that substantially improve this capability. 
All experiments are conducted on data we collected through screenshots of publicly available applications, with requirements automatically generated and subsequently verified by trained human annotators. 
Annotators were compensated fairly at a rate of 1 RMB per data sample, and no personally identifiable or sensitive information is included in the dataset. 
All data that was collected/used contains no information that names or uniquely identifies individual people or offensive content.
We confirm that the benchmark does not introduce bias toward any social group, nor does it involve privacy, security, or harmful use concerns. 
All licenses of these packages allow us for normal research use, and all use of existing artifacts is consistent with their intended use in this paper. 
For the use of AI assistants, we only use AI assistants to polish writing.


% ...

\bibliography{custom}

\begin{thebibliography}{50}
\providecommand{\natexlab}[1]{#1}

\bibitem[{Bai et~al.(2025)Bai, Chen, Liu, Wang, Ge, Song, Dang, Wang, Wang, Tang, Zhong, Zhu, Yang, Li, Wan, Wang, Ding, Fu, Xu, Ye, Zhang, Xie, Cheng, Zhang, Yang, Xu, and Lin}]{Qwen2.5-VL}
Shuai Bai, Keqin Chen, Xuejing Liu, Jialin Wang, Wenbin Ge, Sibo Song, Kai Dang, Peng Wang, Shijie Wang, Jun Tang, Humen Zhong, Yuanzhi Zhu, Mingkun Yang, Zhaohai Li, Jianqiang Wan, Pengfei Wang, Wei Ding, Zheren Fu, Yiheng Xu, and 8 others. 2025.
\newblock \href {https://arxiv.org/abs/2502.13923} {Qwen2.5-vl technical report}.
\newblock \emph{Preprint}, arXiv:2502.13923.

\bibitem[{Bitton et~al.(2023)Bitton, Bansal, Hessel, Shao, Zhu, Awadalla, Gardner, Taori, and Schmidt}]{DBLP:conf/nips/BittonBHSZAGTS23}
Yonatan Bitton, Hritik Bansal, Jack Hessel, Rulin Shao, Wanrong Zhu, Anas Awadalla, Josh Gardner, Rohan Taori, and Ludwig Schmidt. 2023.
\newblock Visit-bench: {A} dynamic benchmark for evaluating instruction-following vision-and-language models.
\newblock In \emph{Advances in Neural Information Processing Systems 36: Annual Conference on Neural Information Processing Systems 2023, NeurIPS 2023, New Orleans, LA, USA, December 10 - 16, 2023}.

\bibitem[{{ByteDance Seed}(2025)}]{Seed1.6}
{ByteDance Seed}. 2025.
\newblock {Seed1.6: Multimodal General-Purpose Model Series}.

\bibitem[{Cheng et~al.(2025)Cheng, Wu, Wu, Ju, Zhang, Zhang, and Liu}]{OS-Kairos}
Pengzhou Cheng, Zheng Wu, Zongru Wu, Tianjie Ju, Aston Zhang, Zhuosheng Zhang, and Gongshen Liu. 2025.
\newblock Os-kairos: Adaptive interaction for mllm-powered {GUI} agents.
\newblock In \emph{Findings of the Association for Computational Linguistics, {ACL} 2025, Vienna, Austria, July 27 - August 1, 2025}, pages 6701--6725. Association for Computational Linguistics.

\bibitem[{DeepSeek-AI(2025)}]{DeepSeek-R1}
DeepSeek-AI. 2025.
\newblock \href {https://arxiv.org/abs/2501.12948} {Deepseek-r1: Incentivizing reasoning capability in llms via reinforcement learning}.
\newblock \emph{Preprint}, arXiv:2501.12948.

\bibitem[{DeepSeek{-}AI et~al.(2025)DeepSeek{-}AI, Guo, Yang, Zhang, Song, Zhang, Xu, Zhu, Ma, Wang, Bi, Zhang, Yu, Wu, Wu, Gou, Shao, Li, Gao, Liu, Xue, Wang, Wu, Feng, Lu, Zhao, Deng, Zhang, Ruan, Dai, Chen, Ji, Li, Lin, Dai, Luo, Hao, Chen, Li, Zhang, Bao, Xu, Wang, Ding, Xin, Gao, Qu, Li, Guo, Li, Wang, Chen, Yuan, Qiu, Li, Cai, Ni, Liang, Chen, Dong, Hu, Gao, Guan, Huang, Yu, Wang, Zhang, Zhao, Wang, Zhang, Xu, Xia, Zhang, Zhang, Tang, Li, Wang, Li, Tian, Huang, Zhang, Wang, Chen, Du, Ge, Zhang, Pan, Wang, Chen, Jin, Chen, Lu, Zhou, Chen, Ye, Wang, Yu, Zhou, Pan, and Li}]{DBLP:journals/corr/abs-2501-12948}
DeepSeek{-}AI, Daya Guo, Dejian Yang, Haowei Zhang, Junxiao Song, Ruoyu Zhang, Runxin Xu, Qihao Zhu, Shirong Ma, Peiyi Wang, Xiao Bi, Xiaokang Zhang, Xingkai Yu, Yu~Wu, Z.~F. Wu, Zhibin Gou, Zhihong Shao, Zhuoshu Li, Ziyi Gao, and 81 others. 2025.
\newblock \href {https://doi.org/10.48550/ARXIV.2501.12948} {Deepseek-r1: Incentivizing reasoning capability in llms via reinforcement learning}.
\newblock \emph{CoRR}, abs/2501.12948.

\bibitem[{Fei et~al.(2025)Fei, Wu, Ji, Zhang, Zhang, Lee, and Hsu}]{DBLP:journals/corr/abs-2501-03230}
Hao Fei, Shengqiong Wu, Wei Ji, Hanwang Zhang, Meishan Zhang, Mong{-}Li Lee, and Wynne Hsu. 2025.
\newblock \href {https://doi.org/10.48550/ARXIV.2501.03230} {Video-of-thought: Step-by-step video reasoning from perception to cognition}.
\newblock \emph{CoRR}, abs/2501.03230.

\bibitem[{Fu et~al.(2023)Fu, Chen, Shen, Qin, Zhang, Lin, Qiu, Lin, Yang, Zheng, Li, Sun, and Ji}]{DBLP:journals/corr/abs-2306-13394}
Chaoyou Fu, Peixian Chen, Yunhang Shen, Yulei Qin, Mengdan Zhang, Xu~Lin, Zhenyu Qiu, Wei Lin, Jinrui Yang, Xiawu Zheng, Ke~Li, Xing Sun, and Rongrong Ji. 2023.
\newblock \href {https://doi.org/10.48550/ARXIV.2306.13394} {{MME:} {A} comprehensive evaluation benchmark for multimodal large language models}.
\newblock \emph{CoRR}, abs/2306.13394.

\bibitem[{Gao et~al.(2024)Gao, Chen, Zhang, Fu, Shen, Zhang, Zhang, Zheng, Sun, Cao, and Ji}]{DBLP:conf/mm/GaoCZFSZZZSCJ24}
Timin Gao, Peixian Chen, Mengdan Zhang, Chaoyou Fu, Yunhang Shen, Yan Zhang, Shengchuan Zhang, Xiawu Zheng, Xing Sun, Liujuan Cao, and Rongrong Ji. 2024.
\newblock \href {https://doi.org/10.1145/3664647.3681249} {Cantor: Inspiring multimodal chain-of-thought of {MLLM}}.
\newblock In \emph{Proceedings of the 32nd {ACM} International Conference on Multimedia, {MM} 2024, Melbourne, VIC, Australia, 28 October 2024 - 1 November 2024}, pages 9096--9105. {ACM}.

\bibitem[{Gemini(2025)}]{Gemini-2.5}
Gemini. 2025.
\newblock \href {https://arxiv.org/abs/2507.06261} {Gemini 2.5: Pushing the frontier with advanced reasoning, multimodality, long context, and next generation agentic capabilities}.
\newblock \emph{Preprint}, arXiv:2507.06261.

\bibitem[{Guo et~al.(2025)Guo, Miao, Wu, Cheng, Zhou, and Zhang}]{OS_GuoYuan}
Yuan Guo, Tingjia Miao, Zheng Wu, Pengzhou Cheng, Ming Zhou, and Zhuosheng Zhang. 2025.
\newblock \href {https://arxiv.org/abs/2506.08972} {Atomic-to-compositional generalization for mobile agents with a new benchmark and scheduling system}.
\newblock \emph{Preprint}, arXiv:2506.08972.

\bibitem[{Hida et~al.(2024)Hida, Ohmura, and Sekiya}]{DBLP:journals/corr/abs-2406-16356}
Rem Hida, Junki Ohmura, and Toshiyuki Sekiya. 2024.
\newblock \href {https://doi.org/10.48550/ARXIV.2406.16356} {Evaluation of instruction-following ability for large language models on story-ending generation}.
\newblock \emph{CoRR}, abs/2406.16356.

\bibitem[{Hu et~al.(2020)Hu, Singh, Darrell, and Rohrbach}]{DBLP:conf/cvpr/HuSDR20}
Ronghang Hu, Amanpreet Singh, Trevor Darrell, and Marcus Rohrbach. 2020.
\newblock \href {https://doi.org/10.1109/CVPR42600.2020.01001} {Iterative answer prediction with pointer-augmented multimodal transformers for textvqa}.
\newblock In \emph{2020 {IEEE/CVF} Conference on Computer Vision and Pattern Recognition, {CVPR} 2020, Seattle, WA, USA, June 13-19, 2020}, pages 9989--9999. Computer Vision Foundation / {IEEE}.

\bibitem[{Ji et~al.(2025)Ji, Chen, Chen, Wu, Qin, and Che}]{MPCC}
Yiyan Ji, Haoran Chen, Qiguang Chen, Chengyue Wu, Libo Qin, and Wanxiang Che. 2025.
\newblock \href {https://arxiv.org/abs/2507.23382} {Mpcc: A novel benchmark for multimodal planning with complex constraints in multimodal large language models}.
\newblock \emph{Preprint}, arXiv:2507.23382.

\bibitem[{Jiang et~al.(2024{\natexlab{a}})Jiang, Wang, Zeng, Zhong, Li, Mi, Shang, Jiang, Liu, and Wang}]{FollowBench}
Yuxin Jiang, Yufei Wang, Xingshan Zeng, Wanjun Zhong, Liangyou Li, Fei Mi, Lifeng Shang, Xin Jiang, Qun Liu, and Wei Wang. 2024{\natexlab{a}}.
\newblock Followbench: A multi-level fine-grained constraints following benchmark for large language models.
\newblock In \emph{ACL (1)}, pages 4667--4688.

\bibitem[{Jiang et~al.(2024{\natexlab{b}})Jiang, Wang, Zeng, Zhong, Li, Mi, Shang, Jiang, Liu, and Wang}]{DBLP:conf/acl/Jiang0ZZLMS00W24}
Yuxin Jiang, Yufei Wang, Xingshan Zeng, Wanjun Zhong, Liangyou Li, Fei Mi, Lifeng Shang, Xin Jiang, Qun Liu, and Wei Wang. 2024{\natexlab{b}}.
\newblock \href {https://doi.org/10.18653/V1/2024.ACL-LONG.257} {Followbench: {A} multi-level fine-grained constraints following benchmark for large language models}.
\newblock In \emph{Proceedings of the 62nd Annual Meeting of the Association for Computational Linguistics (Volume 1: Long Papers), {ACL} 2024, Bangkok, Thailand, August 11-16, 2024}, pages 4667--4688. Association for Computational Linguistics.

\bibitem[{Junlin et~al.(2025)Junlin, Chen, Zhang, and Li}]{junlin2025large}
Xie Junlin, Zhihong Chen, Ruifei Zhang, and Guanbin Li. 2025.
\newblock Large multimodal agents: a survey.
\newblock \emph{Visual Intelligence}, 3(1):24.

\bibitem[{Li et~al.(2024)Li, Bishop, Li, Rawles, Campbell-Ajala, Tyamagundlu, and Riva}]{AndroidControl}
Wei Li, William~E Bishop, Alice Li, Christopher Rawles, Folawiyo Campbell-Ajala, Divya Tyamagundlu, and Oriana Riva. 2024.
\newblock On the effects of data scale on {UI} control agents.
\newblock In \emph{The Thirty-eight Conference on Neural Information Processing Systems Datasets and Benchmarks Track}.

\bibitem[{Li et~al.(2025)Li, Luo, Zhang, Qiu, Huang, and Wei}]{DBLP:conf/naacl/LiLZQHW25}
Zejun Li, Ruipu Luo, Jiwen Zhang, Minghui Qiu, Xuanjing Huang, and Zhongyu Wei. 2025.
\newblock \href {https://doi.org/10.18653/V1/2025.NAACL-LONG.192} {Vocot: Unleashing visually grounded multi-step reasoning in large multi-modal models}.
\newblock In \emph{Proceedings of the 2025 Conference of the Nations of the Americas Chapter of the Association for Computational Linguistics: Human Language Technologies, {NAACL} 2025 - Volume 1: Long Papers, Albuquerque, New Mexico, USA, April 29 - May 4, 2025}, pages 3769--3798. Association for Computational Linguistics.

\bibitem[{Liu et~al.(2025)Liu, Zhao, Liu, Chen, Chai, Ren, Wang, He, and Meng}]{LearnAct}
Guangyi Liu, Pengxiang Zhao, Liang Liu, Zhiming Chen, Yuxiang Chai, Shuai Ren, Hao Wang, Shibo He, and Wenchao Meng. 2025.
\newblock \href {https://arxiv.org/abs/2504.13805} {Learnact: Few-shot mobile gui agent with a unified demonstration benchmark}.
\newblock \emph{Preprint}, arXiv:2504.13805.

\bibitem[{Liu et~al.(2023)Liu, Li, Wu, and Lee}]{LLaVA}
Haotian Liu, Chunyuan Li, Qingyang Wu, and Yong~Jae Lee. 2023.
\newblock Visual instruction tuning.
\newblock In \emph{Advances in Neural Information Processing Systems 36: Annual Conference on Neural Information Processing Systems 2023, NeurIPS 2023, New Orleans, LA, USA, December 10 - 16, 2023}.

\bibitem[{Liu et~al.(2024)Liu, Duan, Zhang, Li, Zhang, Zhao, Yuan, Wang, He, Liu, Chen, and Lin}]{DBLP:conf/eccv/LiuDZLZZYWHLCL24}
Yuan Liu, Haodong Duan, Yuanhan Zhang, Bo~Li, Songyang Zhang, Wangbo Zhao, Yike Yuan, Jiaqi Wang, Conghui He, Ziwei Liu, Kai Chen, and Dahua Lin. 2024.
\newblock \href {https://doi.org/10.1007/978-3-031-72658-3\_13} {Mmbench: Is your multi-modal model an all-around player?}
\newblock In \emph{Computer Vision - {ECCV} 2024 - 18th European Conference, Milan, Italy, September 29-October 4, 2024, Proceedings, Part {VI}}, volume 15064 of \emph{Lecture Notes in Computer Science}, pages 216--233. Springer.

\bibitem[{Lu et~al.(2022)Lu, Mishra, Xia, Qiu, Chang, Zhu, Tafjord, Clark, and Kalyan}]{ScienceQA}
Pan Lu, Swaroop Mishra, Tanglin Xia, Liang Qiu, Kai{-}Wei Chang, Song{-}Chun Zhu, Oyvind Tafjord, Peter Clark, and Ashwin Kalyan. 2022.
\newblock Learn to explain: Multimodal reasoning via thought chains for science question answering.
\newblock In \emph{Advances in Neural Information Processing Systems 35: Annual Conference on Neural Information Processing Systems 2022, NeurIPS 2022, New Orleans, LA, USA, November 28 - December 9, 2022}.

\bibitem[{Luan et~al.(2024)Luan, Feng, Chen, Wang, Zhou, and Li}]{DBLP:journals/corr/abs-2404-09797}
Bozhi Luan, Hao Feng, Hong Chen, Yonghui Wang, Wengang Zhou, and Houqiang Li. 2024.
\newblock \href {https://doi.org/10.48550/ARXIV.2404.09797} {Textcot: Zoom in for enhanced multimodal text-rich image understanding}.
\newblock \emph{CoRR}, abs/2404.09797.

\bibitem[{{Meta AI}(2025)}]{LLaMA-4}
{Meta AI}. 2025.
\newblock {The Llama 4 Herd: The Beginning of a New Era of Natively Multimodal Intelligence}.

\bibitem[{{OpenAI}(2025{\natexlab{a}})}]{GPT-5}
{OpenAI}. 2025{\natexlab{a}}.
\newblock {Introducing GPT-5}.

\bibitem[{{OpenAI}(2025{\natexlab{b}})}]{GPT-o3}
{OpenAI}. 2025{\natexlab{b}}.
\newblock {Introducing OpenAI o3 and o4-mini}.

\bibitem[{Ouyang(2025)}]{DBLP:journals/corr/abs-2504-01805}
Kun Ouyang. 2025.
\newblock \href {https://doi.org/10.48550/ARXIV.2504.01805} {Spatial-r1: Enhancing mllms in video spatial reasoning}.
\newblock \emph{CoRR}, abs/2504.01805.

\bibitem[{Ouyang et~al.(2022)Ouyang, Wu, Jiang, Almeida, Wainwright, Mishkin, Zhang, Agarwal, Slama, Gray, Schulman, Hilton, Kelton, Miller, Simens, Askell, Welinder, Christiano, Leike, and Lowe}]{RLHF}
Long Ouyang, Jeffrey Wu, Xu~Jiang, Diogo Almeida, Carroll Wainwright, Pamela Mishkin, Chong Zhang, Sandhini Agarwal, Katarina Slama, Alex Gray, John Schulman, Jacob Hilton, Fraser Kelton, Luke Miller, Maddie Simens, Amanda Askell, Peter Welinder, Paul Christiano, Jan Leike, and Ryan Lowe. 2022.
\newblock Training language models to follow instructions with human feedback.
\newblock In \emph{Advances in Neural Information Processing Systems}.

\bibitem[{Pan et~al.(2025)Pan, Liu, Wu, Liu, Zhu, Li, Chen, Ouyang, and Rueckert}]{DBLP:journals/corr/abs-2502-19634}
Jiazhen Pan, Che Liu, Junde Wu, Fenglin Liu, Jiayuan Zhu, Hongwei~Bran Li, Chen Chen, Cheng Ouyang, and Daniel Rueckert. 2025.
\newblock \href {https://doi.org/10.48550/ARXIV.2502.19634} {Medvlm-r1: Incentivizing medical reasoning capability of vision-language models (vlms) via reinforcement learning}.
\newblock \emph{CoRR}, abs/2502.19634.

\bibitem[{Peng et~al.(2023)Peng, Li, He, Galley, and Gao}]{SFT_GPT}
Baolin Peng, Chunyuan Li, Pengcheng He, Michel Galley, and Jianfeng Gao. 2023.
\newblock \href {https://arxiv.org/abs/2304.03277} {Instruction tuning with gpt-4}.
\newblock \emph{Preprint}, arXiv:2304.03277.

\bibitem[{Qian et~al.(2025)Qian, Ye, Fauconnier, Grasch, Yang, and Gan}]{DBLP:conf/iclr/QianYFGYG25}
Yusu Qian, Hanrong Ye, Jean{-}Philippe Fauconnier, Peter Grasch, Yinfei Yang, and Zhe Gan. 2025.
\newblock Mia-bench: Towards better instruction following evaluation of multimodal llms.
\newblock In \emph{The Thirteenth International Conference on Learning Representations, {ICLR} 2025, Singapore, April 24-28, 2025}. OpenReview.net.

\bibitem[{Qin et~al.(2024)Qin, Song, Hu, Yao, Cho, Wang, Wu, Liu, Liu, and Yu}]{DBLP:conf/acl/QinSHYCWW00Y24}
Yiwei Qin, Kaiqiang Song, Yebowen Hu, Wenlin Yao, Sangwoo Cho, Xiaoyang Wang, Xuansheng Wu, Fei Liu, Pengfei Liu, and Dong Yu. 2024.
\newblock \href {https://doi.org/10.18653/V1/2024.FINDINGS-ACL.772} {Infobench: Evaluating instruction following ability in large language models}.
\newblock In \emph{Findings of the Association for Computational Linguistics, {ACL} 2024, Bangkok, Thailand and virtual meeting, August 11-16, 2024}, pages 13025--13048. Association for Computational Linguistics.

\bibitem[{Rawles et~al.(2025)Rawles, Clinckemaillie, Chang, Waltz, Lau, Fair, Li, Bishop, Li, Campbell-Ajala, Toyama, Berry, Tyamagundlu, Lillicrap, and Riva}]{AndroidWorld}
Christopher Rawles, Sarah Clinckemaillie, Yifan Chang, Jonathan Waltz, Gabrielle Lau, Marybeth Fair, Alice Li, William~E Bishop, Wei Li, Folawiyo Campbell-Ajala, Daniel~Kenji Toyama, Robert~James Berry, Divya Tyamagundlu, Timothy~P Lillicrap, and Oriana Riva. 2025.
\newblock Androidworld: A dynamic benchmarking environment for autonomous agents.
\newblock In \emph{The Thirteenth International Conference on Learning Representations}.

\bibitem[{Shao et~al.(2024)Shao, Wang, Zhu, Xu, Song, Bi, Zhang, Zhang, Li, Wu, and Guo}]{GRPO}
Zhihong Shao, Peiyi Wang, Qihao Zhu, Runxin Xu, Junxiao Song, Xiao Bi, Haowei Zhang, Mingchuan Zhang, Y.~K. Li, Y.~Wu, and Daya Guo. 2024.
\newblock \href {https://arxiv.org/abs/2402.03300} {Deepseekmath: Pushing the limits of mathematical reasoning in open language models}.
\newblock \emph{Preprint}, arXiv:2402.03300.

\bibitem[{Wan et~al.(2025)Wan, Gao, Mu, Nakov, Wang, and Chen}]{InfoQA}
Kaiyang Wan, Lang Gao, Honglin Mu, Preslav Nakov, Yuxia Wang, and Xiuying Chen. 2025.
\newblock \href {https://doi.org/10.48550/ARXIV.2509.21199} {A fano-style accuracy upper bound for {LLM} single-pass reasoning in multi-hop {QA}}.
\newblock \emph{CoRR}, abs/2509.21199.

\bibitem[{Wang et~al.(2024)Wang, Pan, Shi, Lu, Ren, Zhou, Zhan, and Li}]{MathVision}
Ke~Wang, Junting Pan, Weikang Shi, Zimu Lu, Houxing Ren, Aojun Zhou, Mingjie Zhan, and Hongsheng Li. 2024.
\newblock Measuring multimodal mathematical reasoning with math-vision dataset.
\newblock In \emph{The Thirty-eight Conference on Neural Information Processing Systems Datasets and Benchmarks Track}.

\bibitem[{Wang et~al.(2023)Wang, Kordi, Mishra, Liu, Smith, Khashabi, and Hajishirzi}]{Self-Instruct}
Yizhong Wang, Yeganeh Kordi, Swaroop Mishra, Alisa Liu, Noah~A. Smith, Daniel Khashabi, and Hannaneh Hajishirzi. 2023.
\newblock \href {https://doi.org/10.18653/v1/2023.acl-long.754} {Self-instruct: Aligning language models with self-generated instructions}.
\newblock In \emph{Proceedings of the 61st Annual Meeting of the Association for Computational Linguistics (Volume 1: Long Papers)}, pages 13484--13508, Toronto, Canada. Association for Computational Linguistics.

\bibitem[{Wen et~al.(2024)Wen, Ke, Gu, Wu, Huang, Zhou, Li, Hu, Gao, Xu, Liu, Tang, Wang, and Huang}]{ComplexBench}
Bosi Wen, Pei Ke, Xiaotao Gu, Lindong Wu, Hao Huang, Jinfeng Zhou, Wenchuang Li, Binxin Hu, Wendy Gao, Jiaxing Xu, Yiming Liu, Jie Tang, Hongning Wang, and Minlie Huang. 2024.
\newblock Benchmarking complex instruction-following with multiple constraints composition.
\newblock In \emph{The Thirty-eight Conference on Neural Information Processing Systems Datasets and Benchmarks Track}.

\bibitem[{Wu et~al.(2025)Wu, Wang, Liu, Shi, Yan, Li, Zhu, and Zhang}]{DBLP:conf/acl/WuWLSYLZ025}
Xiaodong Wu, Minhao Wang, Yichen Liu, Xiaoming Shi, He~Yan, Xiangju Li, Junmin Zhu, and Wei Zhang. 2025.
\newblock Lifbench: Evaluating the instruction following performance and stability of large language models in long-context scenarios.
\newblock In \emph{Proceedings of the 63rd Annual Meeting of the Association for Computational Linguistics (Volume 1: Long Papers), {ACL} 2025, Vienna, Austria, July 27 - August 1, 2025}, pages 16445--16468. Association for Computational Linguistics.

\bibitem[{Xiao et~al.(2024)Xiao, Sun, Liu, and Wang}]{LogicVista}
Yijia Xiao, Edward Sun, Tianyu Liu, and Wei Wang. 2024.
\newblock \href {https://doi.org/10.48550/ARXIV.2407.04973} {Logicvista: Multimodal {LLM} logical reasoning benchmark in visual contexts}.
\newblock \emph{CoRR}, abs/2407.04973.

\bibitem[{Xie et~al.(2024{\natexlab{a}})Xie, Zhang, Chen, Zhu, Lou, Tian, Xiao, and Su}]{TravelPlanner}
Jian Xie, Kai Zhang, Jiangjie Chen, Tinghui Zhu, Renze Lou, Yuandong Tian, Yanghua Xiao, and Yu~Su. 2024{\natexlab{a}}.
\newblock Travelplanner: A benchmark for real-world planning with language agents.
\newblock In \emph{Forty-first International Conference on Machine Learning}.

\bibitem[{Xie et~al.(2024{\natexlab{b}})Xie, Zhang, Chen, Li, Zhao, Cao, Hua, Cheng, Shin, Lei, Liu, Xu, Zhou, Savarese, Xiong, Zhong, and Yu}]{OSWorld}
Tianbao Xie, Danyang Zhang, Jixuan Chen, Xiaochuan Li, Siheng Zhao, Ruisheng Cao, Toh~Jing Hua, Zhoujun Cheng, Dongchan Shin, Fangyu Lei, Yitao Liu, Yiheng Xu, Shuyan Zhou, Silvio Savarese, Caiming Xiong, Victor Zhong, and Tao Yu. 2024{\natexlab{b}}.
\newblock Osworld: Benchmarking multimodal agents for open-ended tasks in real computer environments.
\newblock In \emph{Advances in Neural Information Processing Systems 38: Annual Conference on Neural Information Processing Systems 2024, NeurIPS 2024, Vancouver, BC, Canada, December 10 - 15, 2024}.

\bibitem[{Yu et~al.(2024)Yu, Yang, Li, Wang, Lin, Liu, Wang, and Wang}]{DBLP:conf/icml/YuYLWL0WW24}
Weihao Yu, Zhengyuan Yang, Linjie Li, Jianfeng Wang, Kevin Lin, Zicheng Liu, Xinchao Wang, and Lijuan Wang. 2024.
\newblock Mm-vet: Evaluating large multimodal models for integrated capabilities.
\newblock In \emph{Forty-first International Conference on Machine Learning, {ICML} 2024, Vienna, Austria, July 21-27, 2024}. OpenReview.net.

\bibitem[{Yuan et~al.(2023)Yuan, Yuan, Tan, Wang, Huang, and Huang}]{RRHF}
Hongyi Yuan, Zheng Yuan, Chuanqi Tan, Wei Wang, Songfang Huang, and Fei Huang. 2023.
\newblock {RRHF}: Rank responses to align language models with human feedback.
\newblock In \emph{Thirty-seventh Conference on Neural Information Processing Systems}.

\bibitem[{Yue et~al.(2024)Yue, Ni, Zheng, Zhang, Liu, Zhang, Stevens, Jiang, Ren, Sun, Wei, Yu, Yuan, Sun, Yin, Zheng, Yang, Liu, Huang, Sun, Su, and Chen}]{DBLP:conf/cvpr/YueNZ0LZSJRSWYY24}
Xiang Yue, Yuansheng Ni, Tianyu Zheng, Kai Zhang, Ruoqi Liu, Ge~Zhang, Samuel Stevens, Dongfu Jiang, Weiming Ren, Yuxuan Sun, Cong Wei, Botao Yu, Ruibin Yuan, Renliang Sun, Ming Yin, Boyuan Zheng, Zhenzhu Yang, Yibo Liu, Wenhao Huang, and 3 others. 2024.
\newblock \href {https://doi.org/10.1109/CVPR52733.2024.00913} {{MMMU:} {A} massive multi-discipline multimodal understanding and reasoning benchmark for expert {AGI}}.
\newblock In \emph{{IEEE/CVF} Conference on Computer Vision and Pattern Recognition, {CVPR} 2024, Seattle, WA, USA, June 16-22, 2024}, pages 9556--9567. {IEEE}.

\bibitem[{Zhang et~al.(2025{\natexlab{a}})Zhang, Huang, Yao, Liu, Zhang, Lu, and Tao}]{DBLP:journals/corr/abs-2503-12937}
Jingyi Zhang, Jiaxing Huang, Huanjin Yao, Shunyu Liu, Xikun Zhang, Shijian Lu, and Dacheng Tao. 2025{\natexlab{a}}.
\newblock \href {https://doi.org/10.48550/ARXIV.2503.12937} {{R1-VL:} learning to reason with multimodal large language models via step-wise group relative policy optimization}.
\newblock \emph{CoRR}, abs/2503.12937.

\bibitem[{Zhang et~al.(2025{\natexlab{b}})Zhang, Zhu, Shen, Luo, Zhang, Liang, Yang, Lin, Qiao, Chen, Cui, Zhang, and Zhou}]{DBLP:conf/acl/ZhangZSLZLYLQC025}
Tao Zhang, Chenglin Zhu, Yanjun Shen, Wenjing Luo, Yan Zhang, Hao Liang, Fan Yang, Mingan Lin, Yujing Qiao, Weipeng Chen, Bin Cui, Wentao Zhang, and Zenan Zhou. 2025{\natexlab{b}}.
\newblock Cfbench: {A} comprehensive constraints-following benchmark for llms.
\newblock In \emph{Proceedings of the 63rd Annual Meeting of the Association for Computational Linguistics (Volume 1: Long Papers), {ACL} 2025, Vienna, Austria, July 27 - August 1, 2025}, pages 32926--32944. Association for Computational Linguistics.

\bibitem[{Zhang et~al.(2025{\natexlab{c}})Zhang, Zhu, Shen, Luo, Zhang, Liang, Zhang, Yang, Lin, Qiao, Chen, Cui, Zhang, and Zhou}]{CFBench}
Tao Zhang, Chenglin Zhu, Yanjun Shen, Wenjing Luo, Yan Zhang, Hao Liang, Tao Zhang, Fan Yang, Mingan Lin, Yujing Qiao, Weipeng Chen, Bin Cui, Wentao Zhang, and Zenan Zhou. 2025{\natexlab{c}}.
\newblock \href {https://arxiv.org/abs/2408.01122} {Cfbench: A comprehensive constraints-following benchmark for llms}.
\newblock \emph{Preprint}, arXiv:2408.01122.

\bibitem[{Zhang et~al.(2025{\natexlab{d}})Zhang, Wang, Wen, Guo, Zhao, Fang, Ding, Jia, Xiao, Shen et~al.}]{zhang2025large}
Zicheng Zhang, Junying Wang, Farong Wen, Yijin Guo, Xiangyu Zhao, Xinyu Fang, Shengyuan Ding, Ziheng Jia, Jiahao Xiao, Ye~Shen, and 1 others. 2025{\natexlab{d}}.
\newblock Large multimodal models evaluation: a survey.
\newblock \emph{Science China Information Sciences}, 68(12):221301.

\end{thebibliography}


\appendix
\section{Evaluation Prompts}
\label{sec: Evaluation Prompts}
The detailed prompt for evaluating \textsc{FTF-bench} in Section~\ref{sec: benchmark results} is shown below:

\begin{prompt}[title=Prompts for Evaluating \textsc{FTF-bench}]
Given the following task requirements, strictly select the product from the image that best satisfies the criteria. If no product satisfies all the "must-have requirements," explicitly return a refusal. Selection must be based solely on the information provided in the task and visible in the image; no additional reasoning or assumptions are allowed. The rules are as follows:\\

1. Requirement types:\\
- "Must-have requirements" \\
- "Nice-to-have requirements" \\

2. Must-have requirements:\\
- A product must satisfy all must-have requirements simultaneously; missing any requirement disqualifies it.\\
- Any product failing to meet even one must-have requirement is considered ineligible and cannot proceed to further comparison.\\

3. Priority-based selection of nice-to-have requirements (applied only if more than 2 candidate products remain):\\
- Only when multiple products satisfy all must-have requirements should nice-to-have requirements be used for further selection.\\
- Compare products sequentially according to the priority of nice-to-have requirements, from highest to lowest (priority 1 being the highest).\\
- Select the product that satisfies more of the higher-priority requirements. If there is a tie at one priority level, continue comparing the next lower priority until a unique best product is identified.\\
- Example: If priority is ordered as 1>2>3..., and Product A satisfies priorities 1, 2, 3 while Product B satisfies priorities 1, 2, 4, 5, Product A should be chosen because priority 3 is higher than 4 or 5.\\

Output rules:\\
1. If at least one product satisfies all must-have requirements:\\
   - If only one product qualifies, directly output its name.\\
   - If multiple products qualify, apply the nice-to-have priority rules to select the unique best product and output its name.\\
2. If no product satisfies all must-have requirements, output the refusal text: "No product satisfies all must-have requirements."\\

Output format:\\
- Must output strictly in JSON format and only output the JSON.\\
- Format:\\
\{"Answer": ""\}\\
- "Answer": Fill in the name of the selected product. If no product qualifies, set the value to "No product satisfies all must-have requirements" (including quotes).\\

The question is as follows: \\
<question>\\
Strictly follow the above rules and base the selection solely on the information provided in the task and visible in the image.
\end{prompt}

\section{Prompt Design for Benchmark Construction}
\label{sec: Prompt Design for Benchmark Construction}
To systematically construct our benchmark, we designed a set of task-oriented prompts that guide the generation of evaluation questions based on the information present in images. For each task scenario, the prompts explicitly define the core objectives, requirement categories, and output format to ensure consistency and rigor in the generated data. For instance, in the \textit{Single-Answer} scenario, the prompt instructs the model to select one product as the correct answer while providing must-have and nice-to-have requirements with explicit priority levels. Similarly, for \textit{Unanswerable} scenarios, the prompt is carefully constructed to ensure that no product in the image can satisfy all must-have requirements, while optional requirements create a controlled level of confusion. 

\begin{prompt}[title=Prompt for Generating Requirements in Single-Answer Tasks for \textsc{FTF-bench}]
\label{prompt:single-answer}
Given an image displaying products or services, generate a question that asks users to select a product or service based on their needs. Both the question and its answer must be strictly grounded in the information presented in the image; no content should be fabricated. The requirements are as follows:\\

\textbf{1. Determining the correct answer}\\
- Before generating the question, select one product or service from the candidates in the image as the correct answer.\\
- The question must be designed around this selected item, while one or more other items serve as distracting/confusing options.\\
- All content must be strictly based on the image; no invented details are allowed.\\
- \textbf{Core principle}: The question should test the ability to choose the correct item based on user requirements using the visual and contextual information in the image.\\

\textbf{2. Requirement categories and design}\\
- \textbf{Must-have requirements}:\\
  - These requirements must uniquely identify the correct answer, forming a minimal necessary set.\\
  - Each requirement in this set is indispensable—removing any single one may result in multiple items satisfying the criteria.\\
  - The set must not contain redundant requirements—if removing a requirement still uniquely identifies the correct answer, that requirement is considered redundant and should be omitted.\\
- \textbf{Nice-to-have requirements}:\\
  - These are optional requirements that may or may not be satisfied.\\
  - They must be ordered by priority from highest to lowest.\\
  - High-priority nice-to-have requirements are intended to make confusing options more appealing, while the correct answer must satisfy all must-have requirements and may satisfy only a subset of nice-to-have requirements.\\

\textbf{3. Design for confusion}\\
- The question should be challenging and complex.\\
- Distractor options should satisfy more of the high-priority nice-to-have requirements, whereas the correct answer fully satisfies must-have requirements but only partially satisfies nice-to-have requirements.\\
- This encourages users to overemphasize nice-to-have requirements and potentially overlook the must-have requirements, which are the decisive criteria.\\

\textbf{4. Output format}\\
- The output should be a JSON object with the following fields:\\
  - \texttt{"Answer"}: the correct answer\\
  - \texttt{"confusing\_answer"}: the distracting option(s)\\
  - \texttt{"must\_have\_requirements"}: a list of must-have requirements\\
  - \texttt{"nice\_to\_have\_requirements"}: a list of nice-to-have requirements ordered from highest to lowest priority\\
\end{prompt}

\begin{prompt}[title=Prompt for Generating Requirements in Multiple-Answer Tasks for \textsc{FTF-bench}]
\label{prompt:multiple-answer}

Based on the products or services shown in the provided image, generate a question that asks users to select a product or service according to specific requirements. Both the question and the answer must be strictly grounded in the information visible in the image; no content should be fabricated. The requirements are as follows:\\

1. \textbf{Determining the correct answer}\\
- Before generating the question, select one product or service from the image as the correct answer.\\
- All questions must be designed around this selected product, while one or more other products serve as distracting/confusing options.\\
- Core principle: The question should reflect how the correct answer satisfies the requirements relative to the other candidates in the image.\\

2. \textbf{Requirement categories and design}\\
- \textbf{Must-have requirements}:\\
  - These requirements should form a candidate pool containing multiple products from the image, including the correct answer.\\
- \textbf{Nice-to-have requirements}:\\
  - Organize these into a list with explicit priority, from highest to lowest (smaller priority value indicates higher importance).\\
  - By sequentially matching according to priority, the correct answer should be uniquely determined from the candidate pool.\\
- Requirement content should be realistic and contextually appropriate.\\

3. \textbf{Difficulty and distractor design}\\
- The question should be challenging, with complex and diverse requirements.\\
- High-priority nice-to-have requirements may conflict with must-have requirements to increase difficulty.\\
- Distractor options should satisfy more of the high-priority nice-to-have requirements but fail to meet must-have requirements; the correct answer may satisfy fewer high-priority nice-to-have requirements but must fully satisfy all must-have requirements.\\

4. \textbf{Output format}\\
- The output must be strictly in JSON format and contain only the JSON object, with the following fields:\\
  - \texttt{"mandatory\_requirements"}: a list of must-have requirements\\
  - \texttt{"optional\_requirements"}: a list of nice-to-have requirements with priorities, each represented as an object containing \texttt{"priority"} and \texttt{"requirement"}\\
  - \texttt{"confusing\_answer"}: the distracting option\\
  - \texttt{"final\_selected\_product"}: the product ultimately selected after applying the optional requirements\\
  
Strictly follow the above rules and base the selection solely on the information provided in the task and visible in the image.
\end{prompt}

\begin{prompt}[title=Prompt for Generating Requirements in Unanswerable Tasks for \textsc{FTF-bench}]
\label{prompt:unanswerable}
Based on the products or services shown in the provided image, generate a question that asks users to select a product or service according to specific requirements. The question must be designed so that no product or service in the image can satisfy all core requirements. Both the question and its answer must be strictly grounded in the information visible in the image; no content should be fabricated. The requirements are as follows:\\

1. \textbf{Core objective}\\
- The generated question must ensure that all products or services in the image cannot become the correct answer; that is, no product/service can fully satisfy the task’s core requirements.\\

2. \textbf{Requirement categories and design}\\
- \textbf{Must-have requirements}:\\
  - Must satisfy two conditions:\\
    1. Each individual requirement should be satisfied by at least one product or service in the image.\\
    2. When all must-have requirements are combined, no single product or service should satisfy them all (i.e., each requirement has a corresponding “matching product,” but no product matches all requirements).\\
- \textbf{Nice-to-have requirements (optional/distractor requirements)}:\\
  - Organize as a list with explicit priority, from highest to lowest (smaller priority value indicates higher importance).\\
  - Some optional requirements should be satisfied by most candidate products to create the illusion that a perfect product exists.\\
- Each requirement should be fluent, reasonable, and reflect common, real-world user needs.\\

3. \textbf{Difficulty and distractor design}\\
- The question should be complex and challenging, emphasizing confusion.\\
- Each product/service in the image should satisfy most of the must-have requirements, failing only on 1–2 critical requirements.\\
- Each product/service should satisfy some high-priority nice-to-have requirements, creating a “each has advantages” scenario, making it difficult to realize that no product fully satisfies all must-have requirements.\\

4. \textbf{Output format}\\
- The output must be strictly in JSON format and contain only the JSON object, with the following fields:\\
  - \texttt{"mandatory\_requirements"}: list of selected must-have requirements\\
  - \texttt{"optional\_requirements"}: list of selected nice-to-have requirements with priority, each represented as an object containing \texttt{"priority"} and \texttt{"requirement"}\\
  - \texttt{"candidate\_products"}: list of products that satisfy some, but not all, must-have requirements (serving as candidate options)\\

Strictly follow the above rules and base the selection solely on the information provided in the task and visible in the image. Ensure that no single product fully satisfies all must-have requirements.
\end{prompt}

\section{Details of Human Verification}
\label{sec: Details of Human Verification}
\begin{figure*}[t!]
  \centering
  \includegraphics[width=0.98\textwidth]{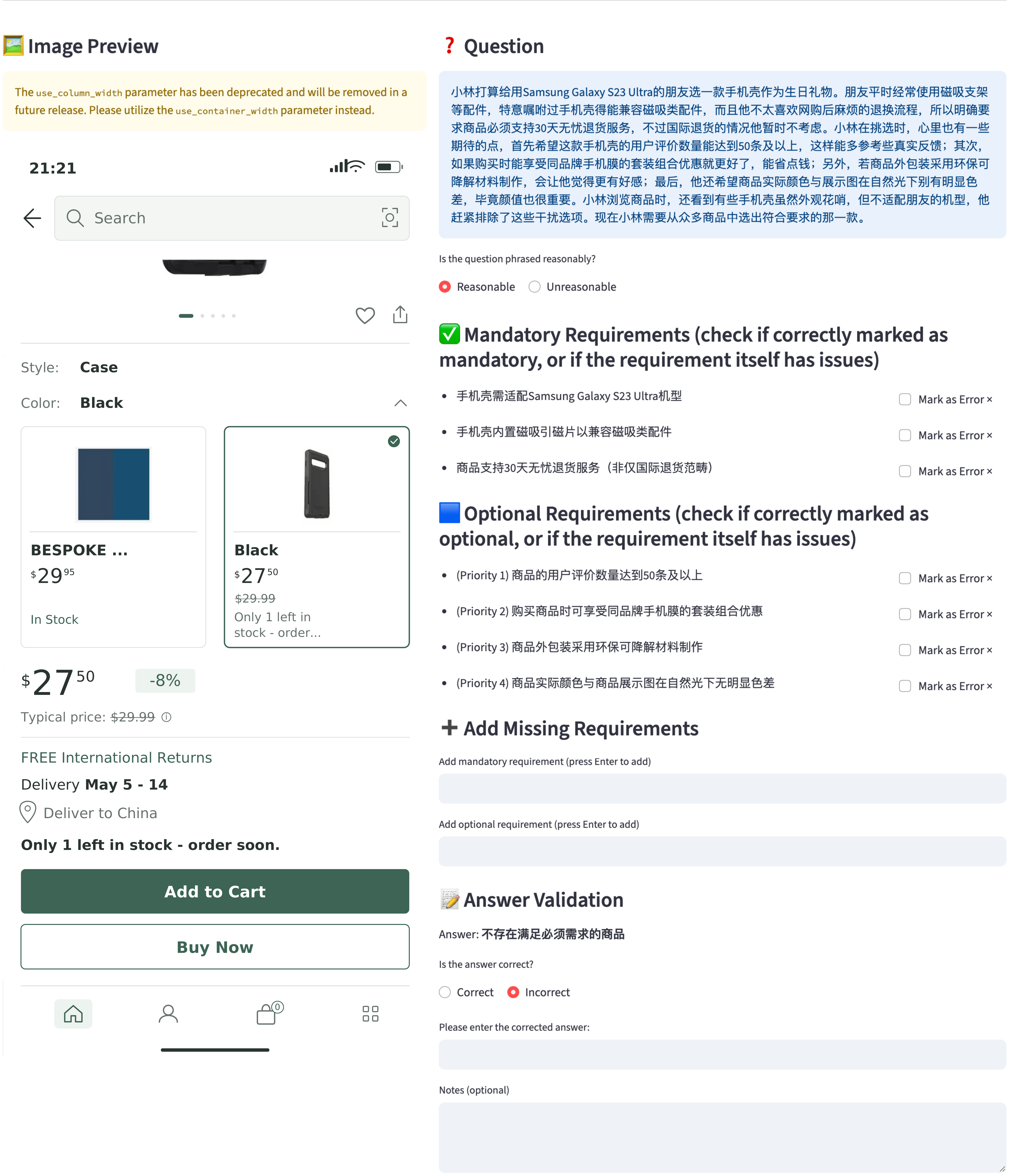}
  \caption{Screenshot of the human verification annotation interface. Annotators are asked to evaluate question reasonableness, answer accuracy, requirement categorization, and description correctness.}
  \label{fig: human verification}
\end{figure*}

To ensure the quality and reliability of our dataset, we implemented a rigorous human verification process. Specifically, we developed a custom annotation script that streamlined the verification workflow, as shown
in figure \ref{fig: human verification}, allowing annotators to efficiently inspect each instance. The script presents the input data, associated candidate options, and relevant contextual information in a user-friendly interface, while logging the annotators' selections and comments for further analysis.

Annotators were instructed to evaluate multiple aspects of each item. These include the reasonableness of the question, the accuracy of the answer, the correctness of requirement categorization and requirement descriptions. Each instance was independently reviewed by at least two annotators. Disagreements were resolved through discussion or adjudication by a senior annotator. This procedure not only guarantees high annotation accuracy but also establishes a transparent and reproducible verification pipeline.

\begin{figure}[t!]
  \centering
  \includegraphics[width=0.36\textwidth]{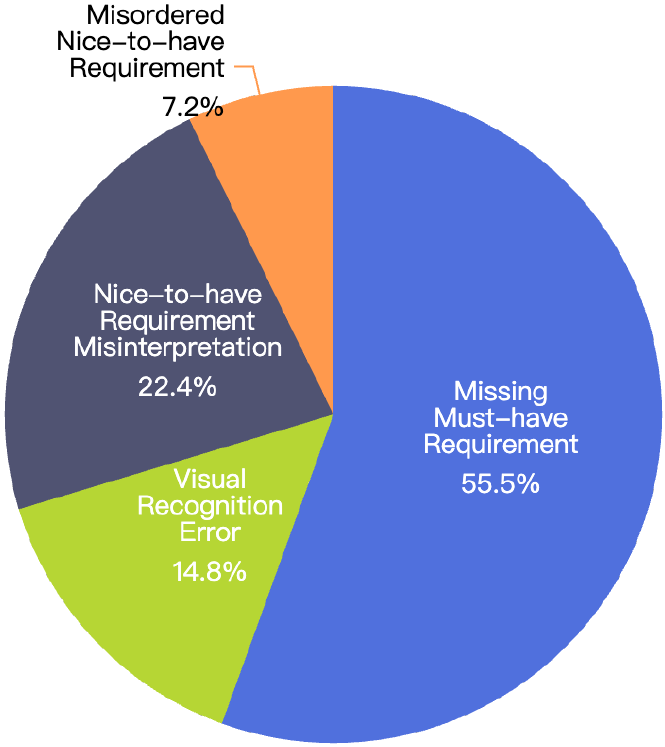}
  \caption{Distribution of model error types identified during human verification. Models frequently failed to follow \textit{must-have requirements}, sometimes treated \textit{nice-to-have requirements} as mandatory, and occasionally misordered them, highlighting challenges in understanding and prioritizing user preferences.}
  \label{fig: wrong reason}
\end{figure}

During the human verification process, we systematically examined the model's erroneous predictions and identified several prominent error types, as shown in Figure \ref{fig: wrong reason}. The model often failed to follow \textit{must-have requirements},  and sometimes treated \textit{nice-to-have requirements} as mandatory and occasionally misordered them, reflecting challenges in understanding and prioritizing user preferences.

\section{Model-Dependent Bias in Data Creation and Evaluation}
\label{app:model-bias}

In our pipeline, we use \texttt{Doubao-Seed-1.6-250615} to generate the initial requirement sets and candidate answers, and also as the judger. 
All items are then checked by human annotators, who verify both the realism of the user requirements and the correctness of the final answer. 
Table~\ref{tab:human-judgment-distribution} summarizes the joint distribution of human judgments on question reasonableness and answer correctness.

\begin{table*}[t!]
    \centering
    \small
    \resizebox{0.8\linewidth}{!}{
    \begin{tabular}{lccc}
        \toprule
        {\textbf{Human judgment}} & {\textbf{Answer has error}} & {\textbf{Answer correct}} & {\textbf{Total}} \\
        \midrule
        {Question unreasonable} & {18.72\%} & {1.41\%} & {20.13\%} \\
        {Question reasonable} & {9.90\%} & {69.98\%} & {79.87\%} \\
        \midrule
        {Total} & {28.62\%} & {71.38\%} & {100\%} \\
        \bottomrule
    \end{tabular}}
    \caption{{Joint distribution of human judgments on question reasonableness (question\_reasonable) and answer correctness (has\_error).}}
    \label{tab:human-judgment-distribution}
\end{table*}

Most discrepancies come from prompts that annotators consider unrealistic in real life, while the majority of retained items have both reasonable requirements and correct answers. This indicates that \texttt{Doubao-Seed-1.6-250615} is capable of serving as a generator/judger, and that the subsequent human pass substantially reduces residual errors and potential same-source bias in the benchmark.

\section{Training Dynamics of \textsc{FTF-rl} on Qwen2.5-VL-7B-Instruct}
\label{app:ftf-rl-dynamics}
We further analyze the behavior of \textsc{FTF-rl} by tracking the evolution of different rewards over the whole training process on Qwen2.5-VL-7B-Instruct. 
We monitor the format reward, the requirement classification reward, the answer correctness reward, and the aggregated overall reward at multiple training steps.

\begin{figure*}[t!]
    \centering
    \includegraphics[width=0.9\linewidth]{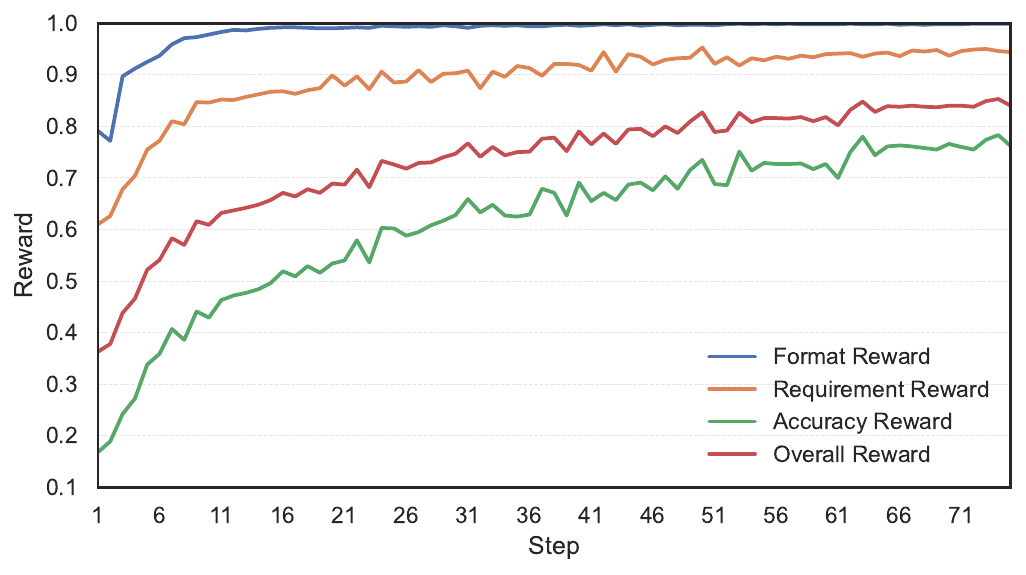}
    \caption{
    {Training reward curves of Qwen2.5-VL-7B-Instruct under \textsc{FTF-rl}.}}
    \label{fig:ftf-rl-training-curves}
\end{figure*}

As shown in Figure~\ref{fig:ftf-rl-training-curves}, all four rewards improve monotonically during training, leading to a steady rise in the overall reward.

\section{Qualitative Case Studies on General Reasoning Benchmarks}
\label{app:qualitative_reasoning}

To illustrate how requirement-aware reasoning learned from \textsc{FTF-bench} transfers to other tasks, we present two short case studies on LogicVista and MathVision using Qwen2.5-VL-7B-Instruct trained with \textsc{FTF-rl}. 
In both examples, the model only sees the original benchmark input and no additional supervision.

We provide the two cases in Table~\ref{tab:mathvision_case}~and~\ref{tab:logicvista_case}. 
In both cases, the trained model follows a consistent pattern. 
The first case is a MathVision geometry problem about a circular carpet on a tiled floor. 
The model must decide which grey tile pattern cannot come from any circle. After \textsc{FTF-rl}, Qwen2.5-VL-7B-Instruct does not jump directly to an option. 
It first summarizes the task requirement that grey tiles must be exactly those intersected by a single convex circle, so they should form one connected region with a smooth boundary. 
It then explicitly plans to check each candidate against this requirement and finally concludes that the option with a disconnected grey region is impossible. 
This shows that the model uses a requirement-driven plan rather than local pattern matching.

The second case is a LogicVista problem that involves inferring the meaning of two symbolic operations applied to shapes. 
The model must fill in a missing output shape and a missing operation symbol. 
After \textsc{FTF-rl}, Qwen2.5-VL-7B-Instruct begins by restating the subgoals, then applies these inferred rules to the two queries. 
It then follows this plan step by step and correctly selects the option. The model learns to organize the task into requirement extraction and execution.

\begin{table*}[t!]
\centering
\begin{tabularx}{\textwidth}{@{}X@{}}
\toprule

\textbf{Question.}
A circular carpet is placed on a floor which is covered by equally big, square tiles. All tiles that have at least one point in common with the carpet are coloured in grey. Which of the following cannot be a result of this?
<image> 
\\
\addlinespace[0.5em]
{\textbf{Image.}}\\[0.2em]
\includegraphics[width=0.8\linewidth]{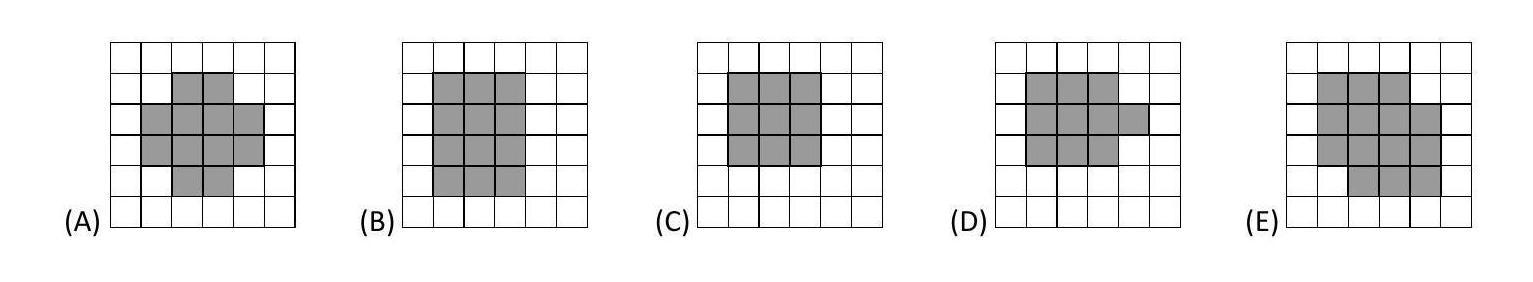} \\
\addlinespace[0.5em]
\begin{minipage}[t]{\linewidth}

\textbf{Output.}\\
<requirements>{"mandatory": ["Identify which grey tile pattern CANNOT result from coloring all tiles touching a circular carpet (tiles with at least one point in common with the circle)"],"optional": ["Explain why the pattern is invalid"]}</requirements>\\
<think>\\
To solve this problem, we need to understand how placing a circular carpet would affect the surrounding squares if they touch it anywhere. \\
A circle is a convex, smooth, and symmetric shape. Tiles colored by a circle (all tiles intersecting the circle) must form a connected, "rounded" boundary (no sharp corners or disconnected "extensions" that cannot be touched by a circle).\\
Let’s analyze each option:\\
A: The shape is connected with smooth edges. A circle could touch these tiles.\\
B: A tall, narrow shape—still connected and smooth; a circle could align with this.\\
C: A compact, connected shape; a circle could fit this.\\
D: The shape has a small extra tile on the right side but remains connected and smooth. A circle could touch this.\\
E: The the bottom-right tiles are separated from the main block by an empty tile. A circle is convex and continuous—it cannot touch tiles that are disconnected from the main group. The empty tile between the main block and the lower-right tiles means the circle cannot reach both regions without also touching the empty tile.\\
Thus, E cannot be the result.\\
</think>\\
<answer>E</answer>

\end{minipage}
 \\
\bottomrule
\end{tabularx}
\caption{{Qualitative case study on MathVision after training on Qwen2.5-VL-7B-Instruct.}}
\label{tab:mathvision_case}
\end{table*}

\begin{table*}[t!]
\centering
\begin{tabularx}{\textwidth}{@{}X@{}}
\toprule

\textbf{Question.}
Based on the diagram at the top of the page, which two of the proposals A, B, C or D completes the diagrams at the bottom of the page?
<image> 
\\
\addlinespace[0.5em]
{\textbf{Image.}}\\[0.2em]
\includegraphics[width=0.8\linewidth]{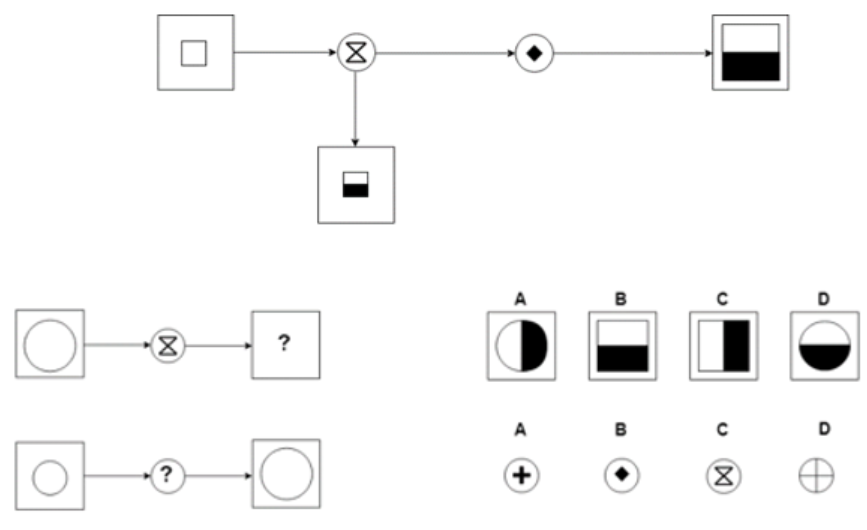} \\
\addlinespace[0.5em]
\begin{minipage}[t]{\linewidth}

\textbf{Output.}\\
<requirements>{"mandatory": ["Analyze the top diagram to determine the transformation rules of operations for shapes within square frames","Apply the rules to select the correct shape (from A/B/C/D) for the '?' in the first bottom diagram","Apply the rules to select the correct operation (from A/B/C/D) for the '?' in the second bottom diagram"],"optional": ["Explain each operation's function with reference to the top diagram"]}</requirements>\\
<think>\\
First, I need to re-extract the operation rules from the top diagram. The top diagram has two inputs: a square with a small white square, and a square with a small square.\\
Operation cross symbol combines the "small inner shapes" of the two inputs.\\
Operation the diamond symbol then scales the combined small shape to fill the entire outer square frame.\\
Next, for the first bottom diagram: Input is a square with a small white circle.\\
Operation cross symbol should combine this small circle with the "bottom half black" fill.\\
So the result should be a square frame containing a small circle (bottom half black) — which is option D.\\
For the second bottom diagram:\\
Input is a square with a small white circle; output is a square with a large circle.\\
This scaling function matches operation the diamond symbol from the top diagram — which is option B.\\
</think>\\
<answer>D, B</answer>

\end{minipage}
 \\
\bottomrule
\end{tabularx}
\caption{{Qualitative case study on LogicVista after training on Qwen2.5-VL-7B-Instruct.}}
\label{tab:logicvista_case}
\end{table*}

\section{Additional Experiments on Complex Agent Tasks}
\label{sec:appendix-agent}

To further assess the generality of requirement-aware reasoning beyond the three service-oriented domains in our main experiments, we conduct additional evaluations on complex agent tasks. We select two representative benchmarks.

AndroidControl~\citep{AndroidControl} evaluates mobile GUI control in realistic Android environments. 
It consists of human demonstrations of everyday tasks across diverse apps, where each trajectory records the screen observations, natural language instructions, and the corresponding low-level actions executed by the user. 
We evaluate performance on AndroidControl using Task Match Rate (TMR) and Action Match Rate (AMR) as our main metrics.

ScienceQA~\citep{ScienceQA} captures multi-step scientific reasoning, which serves as a comprehensive testbed for multimodal reasoning across diverse science topics. 
It features a diverse collection of science questions covering 26 topics and 127 categories, where each example is annotated with CoT explanations and lectures to assess the model's ability to perform complex scientific reasoning and explanation generation. 
The metric is accuracy, measuring the number of corrected answers provided by models.


Table~\ref{tab:agent-generalization}~and~\ref{tab:gui_metrics} report the performance before and after \textsc{FTF-rl} training on ScienceQA and AndroidControl, respectively. Across both benchmarks, we observe consistent improvements after \textsc{FTF-rl}, mirroring the gains we reported previously on LogicVista, MathVision, and MathVista. 
This shows that learning to prioritize requirements on \textsc{FTF-bench} transfers to more complex agent behaviors.

\begin{table*}[t!]
    \centering
    \resizebox{0.4\linewidth}{!}{
    \begin{tabular}{lccc}
        \toprule
        {Model} &  {ScienceQA} \\
        \midrule
        {Qwen2.5-VL-7B-Instruct}  & {40.06} \\
        {$\quad$+\textsc{FTF-rl}}  & {\textbf{40.28}\positive{0.22}} \\
        \bottomrule
    \end{tabular}}
    \caption{{Performance on ScienceQA after training with \textsc{FTF-rl} on \textsc{FTF-bench}.}}
    \label{tab:agent-generalization}
\end{table*}

\begin{table*}[t!]
    \centering
    \resizebox{1.0\linewidth}{!}{
    \begin{tabular}{ll|ccccccc}
        \toprule
        {Model} & {Metric} & {CLICK} & {TYPE} & {SCROLL} & {OPENAPP} & {WAIT} & {COMPLETE} & {PRESS} \\
        \midrule
        \multirow{2}{*}{{Qwen2.5-VL-7B-Instruct}} & {TMR} & {0.9538} & {0.8880} & {0.8571} & {0.0000} & {0.0459} & {0.9339} & {0.2857} \\
         & {AMR} & {0.3317} & {0.6800} & {0.0603} & {0.0000} & {0.0459} & {0.9339} & {0.2857} \\
        \midrule
        \multirow{2}{*}{{$\quad$+\textsc{FTF-rl}}} & {TMR} & {0.9608\positive{0.0071}} & {0.9216\positive{0.0336}} & {0.8026\negative{0.0545}} & {0.0000\positive{0.0000}} & {0.0529\positive{0.0071}} & {0.9533\positive{0.0194}} & {0.5364\positive{0.2507}} \\
         & {AMR} & {0.3240\negative{0.0077}} & {0.7184\positive{0.0384}} & {0.0446\negative{0.0157}} & {0.0000\positive{0.0000}} & {0.0529\positive{0.0071}} & {0.9533\positive{0.0194}} & {0.5364\positive{0.2507}} \\
        \bottomrule
    \end{tabular}}
    \caption{{Performance on AndroidControl after training with \textsc{FTF-rl} on \textsc{FTF-bench}.}}
    \label{tab:gui_metrics}
\end{table*}

\section{Analysis of the 10\% Evaluation Subset}
\label{app:subset-analysis}

In the RL experiments, we randomly sample 90\% of FTF-Bench for training and hold out the remaining 10\% for evaluation, and the same subset is used for all models. 
To check whether this subset introduces bias, we evaluate all baseline models on both the full benchmark and this 10\% subset and compare the results in Table~\ref{tab:10_eval}.

\begin{table*}[t!]
    \centering
    \resizebox{1.0\linewidth}{!}{
    \begin{tabular}{ll|cccccccc}
        \toprule
        {Model} & {Split} & {Sin. Upper} & {Sin. Direct} & {Mul. Upper} & {Mul. Direct} & {Unans. Upper} & {Unans. Direct} & {Avg. Upper} & {Avg. Direct} \\
        \midrule
        \multirow{2}{*}{{Gemini-2.5-pro}} & {Full}   & {88.89} & {86.91} & {84.20} & {82.26} & {81.72} & {78.55} & {84.26} & {81.75} \\
        & {Subset} & {89.47} & {88.36} & {85.13} & {83.74} & {83.09} & {80.21} & {86.44} & {84.91} \\
        \multirow{2}{*}{{GPT-o3}}           & {Full}   & {82.67} & {77.78} & {80.41} & {80.03} & {83.31} & {83.09} & {82.33} & {79.68} \\
        & {Subset} & {84.18} & {79.42} & {82.07} & {81.36} & {85.27} & {84.16} & {83.41} & {81.09} \\
        \multirow{2}{*}{{Qwen2.5-VL-7B-Instruct}}  & {Full}   & {61.16} & {22.99} & {58.25} & {20.10} & {47.59} & {18.64} & {57.69} & {21.05} \\
         & {Subset} & {66.83} & {46.38} & {63.17} & {32.52} & {52.74} & {44.26}  & {62.81} & {39.78} \\
        \multirow{2}{*}{{Qwen2.5-VL-32B-Instruct}} & {Full}   & {70.66} & {69.25} & {70.55} & {70.17} & {53.08} & {50.58} & {67.72} & {66.57} \\
        & {Subset} & {72.11} & {71.04} & {73.26} & {72.18} & {54.79} & {58.42} & {69.71} & {69.43} \\
        \bottomrule
    \end{tabular}}
    \caption{{Performance of all models on the full \textsc{FTF-bench} and on the 10\% evaluation subset.}}
    \label{tab:10_eval}
\end{table*}

\section{Additional Analysis on Decoding Temperature}
\label{sec:temp_analysis}

We ran an additional experiment on \textsc{Doubao-1.6-seed} with different decoding temperatures, including a fully deterministic setting with temperature $0.0$. 
Table~\ref{tab:temp-accuracy} reports the accuracy on \textsc{FTF-bench} for temperatures between $0.0$ and $1.0$.

\begin{table*}[t!]
    \centering
    \begin{tabular}{lcccc}
        \toprule
        {Temperature} & {Single-Answer} & {Multiple-Answer} & {Unanswerable} & {Average} \\
        \midrule
        {0.0} & {78.66} & {76.99} & {84.05} & {80.59} \\
        {0.2} & {88.16} & {80.57} & {73.61} & {81.70} \\
        {0.4} & {79.41} & {80.96} & {62.28} & {72.21} \\
        {0.6} & {81.10} & {79.10} & {61.37} & {71.92} \\
        {0.8} & {90.24} & {82.89} & {77.55} & {82.65} \\
        {1.0} & {89.47} & {82.29} & {77.81} & {82.38} \\
        \bottomrule
    \end{tabular}
    \caption{{\textsc{Doubao-1.6-seed} accuracy on \textsc{FTF-bench} under different decoding temperatures.}}
    \label{tab:temp-accuracy}
\end{table*}

The overall accuracy varies only moderately across temperatures, indicating that the tendency to refuse is stable.

We further measure the agreement between outputs at temperature 0 and outputs at other temperatures on an instance-by-instance basis.
Results are shown in Table~\ref{tab:temp-agreement}.

\begin{table*}[t!]
    \centering
    \begin{tabular}{lcccc}
        \toprule
        {Temperature} & {Single-Answer} & {Multiple-Answer} & {Unanswerable} & {Average} \\
        \midrule
        {0.2} & {79.19} & {75.24} & {80.89} & {77.79} \\
        {0.4} & {74.00} & {73.33} & {67.16} & {70.72} \\
        {0.6} & {73.55} & {73.47} & {66.55} & {70.47} \\
        {0.8} & {79.11} & {75.22} & {78.34} & {77.74} \\
        {1.0} & {78.84} & {75.54} & {79.21} & {78.13} \\
        \bottomrule
    \end{tabular}
    \caption{{Instance-level agreement between temperature $0.0$ and higher temperatures for \textsc{Doubao-1.6-seed}.}}
    \label{tab:temp-agreement}
\end{table*}

This high consistency shows that the refusal behavior and the error patterns we analyze are stable.

\end{document}